\documentclass[a4paper,fleqn]{cas-sc}

\usepackage[numbers]{natbib}
\usepackage{algorithm}
\usepackage{amssymb}
\usepackage{lipsum}
\usepackage{algpseudocode}
\usepackage{amsthm}
\usepackage{amsmath}
\usepackage{graphicx}
\usepackage{caption}
\usepackage{subfig} 
\usepackage{tabularx}
\usepackage{booktabs}
\usepackage{array}
\usepackage{float}
\usepackage{color}
\usepackage{graphicx}
\usepackage{multirow}
\usepackage{booktabs}
\usepackage{placeins}
\usepackage{subfig}
\usepackage{adjustbox}
\usepackage{float} 
\usepackage{placeins} 
\def\tsc#1{\csdef{#1}{\textsc{\lowercase{#1}}\xspace}}
\tsc{WGM}
\tsc{QE}
\tsc{EP}
\tsc{PMS}
\tsc{BEC}
\tsc{DE}
\newtheorem{example}{Example}[section]

\begin{document}
\let\WriteBookmarks\relax
\def\floatpagepagefraction{1}
\def\textpagefraction{.001}

\shorttitle{Adaptive Deep Learning for Breast Cancer Subtype Prediction via Misprediction Risk Analysis}

\shortauthors{Gul Sheeraz et~al.}

\title [mode = title]{Adaptive Deep Learning for Breast Cancer Subtype Prediction Via Misprediction Risk Analysis}                      

\author[1]{Gul Sheeraz}[type=author, 
                        orcid=0000-0002-8885-7601, 
]

\author[2]{Liu Feiyu}[type=author]
\author[1]{Chen Qun}[type=author]
\author[3]{Zhou Fengjin}[type=author]

\cormark[1] 
\ead{sheerazgul@mail.nwpu.edu.cn}
\ead{1015448993@qq.com}
\ead{chenbenben@nwpu.edu.cn}
\ead{dr.zhoufj@xjtu.edu.cn}


\affiliation[1]{organization={School of Computer Science, Northwestern Polytechnical University},
            city={Xian},
            state={Shaanxi},
            country={China}}
\affiliation[2]{organization={School of Software, Northwestern Polytechnical University},
            city={Xian},
            state={Shaanxi},
            country={China}}

\affiliation[3]{organization={Honghui Hospital, Xian Jiao Tong University},
            city={Xian},
            state={Shaanxi},
            country={China}}

\cortext[cor1]{Corresponding author: dr.zhoufj@xjtu.edu.cn}

\begin{abstract}
Breast cancer remains a leading cause of cancer-related mortality worldwide. Early detection is critical for improving patient outcomes, yet manual histopathology analysis is complex and subject to inter-observer variability. While computer-aided diagnostic systems based on deep neural networks have advanced binary prediction tasks, they struggle with multiclass subtype prediction due to inter-class similarity, class imbalance, and domain shifts, resulting in frequent mispredictions. This study aims to develop an adaptive learning framework that quantifies and mitigates misprediction risk of the neural networks in breast cancer subtype prediction from histopathology images. We propose MultiRisk, a multiclass misprediction risk analysis model that ranks the likelihood of misprediction for each image using interpretable features derived from heterogeneous deep neural network (DNN) representations, with a dedicated risk model constructed and trained to capture multiclass risk patterns. Building on this, we introduce a risk-based adaptive learning strategy that fine-tunes prediction models based on dataset-specific characteristics, effectively reducing misprediction risk and improving adaptability to diverse workloads to support clinical decision systems. The MultiRisk framework is evaluated on multiple histopathological image datasets. For misprediction risk analysis, it achieves AUROCs of 78.1\%, 75.6\%, and 76.3\%, achieving benchmark AUROC performance. Risk-based adaptive training further improves F1-scores to 61.15\%, 65.98\%, and 80.53\%, achieving benchmark F1 performance and demonstrating the effectiveness of both risk analysis and adaptive learning across resolutions and domain shifts. MultiRisk provides an interpretable, adaptive, and effective framework for breast cancer subtype prediction. By combining misprediction risk analysis with adaptive fine-tuning, it improves predictive accuracy of the neural networks, mitigates errors under limited labeled data, and generalizes across domains, different cancer types, and various model architectures, supporting reliable clinical decision-making. Our code is available at: \href{https://github.com/SheerazNWPU/MultiRisk}{https://github.com/SheerazNWPU/MultiRisk}

\end{abstract}


\maketitle

\section{Introduction}
Breast cancer, with over 2.3 million cases annually, stands as the most prevalent adult cancer, and globally, it is the leading or second leading cause of cancer-related deaths in 95\% of countries~\cite{1,2}. Timely detection is crucial to reducing mortality rates in breast cancer-related deaths. The detection of breast cancer, a heterogeneous collection of diseases arising from epithelial tissue abnormalities, requires various radiological approaches (e.g., Mammography, Ultrasound, MRI)~\cite{3, 4}. However, these methods may not fully capture heterogeneity. Histopathological studies, including biopsy and microscopic examination, offer in-depth information on breast tissue characteristics~\cite{5, 6}.


Histopathological images of breast tumors contain crucial information such as tumor type and cancer stage. However, manual analysis by experienced pathologists is time-consuming and prone to errors. A study indicates a notable misdiagnosis rate, with 1 in 71 biopsies and 1 in 5 cancer cases being misclassified~\cite{7}. Table~\ref{tab:pathologist_f1} shows the weighted F1-scores from a study~\cite{33} of domain expert pathologists compared to one of our trained baseline DenseNet-121 for 7-class breast cancer sub-typing on the BRACS dataset, highlighting the difficulty of accurate subtype prediction even for experienced pathologists and the potential of automated approaches to assist in accurate diagnosis and subtype prediction.

\begin{table}[h!]
\centering
\caption{Weighted F1-scores comparison between pathologists and a baseline DenseNet-121 for 7-class breast cancer subtyping on the BRACS dataset~\cite{33}.}
\label{tab:pathologist_f1}
\begin{tabular}{lcc}
\toprule
 & Weighted F1 (\%) \\
\midrule
Pathologist 1 & 55.30 \\
Pathologist 2 & 57.07 \\
Pathologist 3 & 56.71 \\
Pathologist statistics (mean ± std) & 56.36 ± 0.76 \\
DenseNet-121 statistics (mean ± std) & 60.65 ± 4.2 \\
\bottomrule
\end{tabular}
\end{table}

Computer-Aided Diagnosis aids in accurate diagnosis, especially in binary prediction~\cite{8}. Recent advances in AI have introduced state-of-the-art approaches~\cite{11,12,13,14,16}, predominantly using deep learning and convolutional neural networks (CNNs) for optimal performance in binary and subtype prediction. However, even these DNN models face several critical challenges in practical scenarios. First, the limited availability of labeled histopathological images often leads to insufficient representation of rare subtypes, causing the neural networks poor generalization. Second, data distribution shifts between training and deployment settings, such as variations in staining, slide preparation, and patient demographics, can significantly degrade deep models performance. Third, the subtype prediction is inherently more challenging due to high inter-class similarity and intra-class heterogeneity, which results in frequent mispredictions even for well-trained models. Consequently, a model that performs well on the training data may still yield unreliable predictions on new samples, limiting its clinical applicability. 

\begin{figure*} 
  \centering
  \includegraphics[width=\textwidth]{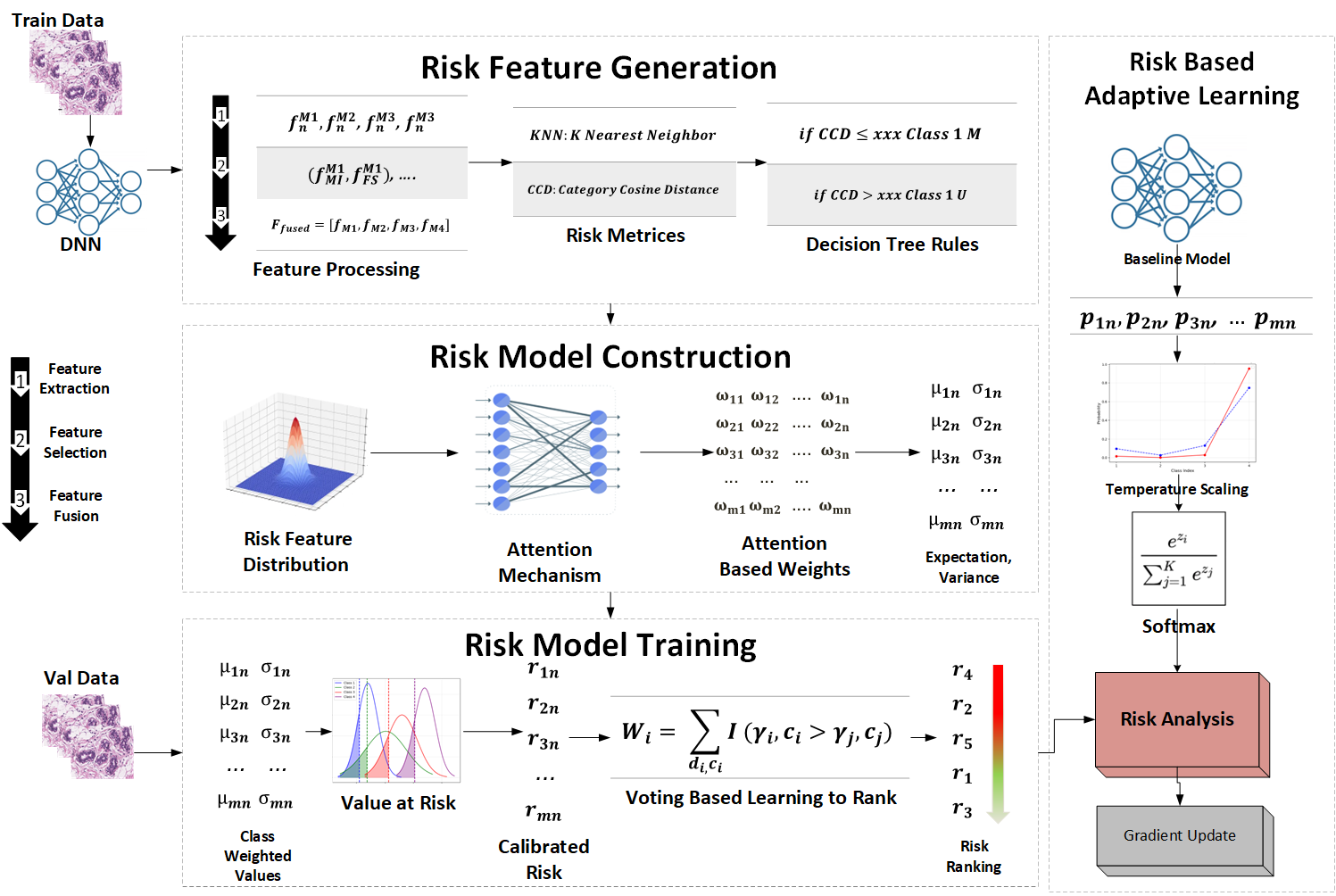}
  \caption{Misprediction Risk Analysis for Breast Cancer Subtype Prediction}
  \label{fig:risk_analysis}
\end{figure*}

To address these challenges, we propose an adaptive deep learning framework for breast cancer subtype prediction from H\&E-stained histopathological images via multiclass misprediction risk analysis. We first present a novel solution for multiclass risk analysis that can effectively quantify a sample's misprediction risk by a DNN model. As shown in Figure~\ref{fig:risk_analysis}, our approach advances beyond the binary formulation of LearnRisk. Our proposed solution, MultiRisk, consists of novel risk feature generation for multiclass settings, attention-based risk model construction and optimized training for multiclass scenarios, and a risk-based adaptive training approach for breast cancer subtype prediction.

Risk feature generation employs a fusion-based strategy that combines heterogeneous features through feature selection, diverse risk metric extraction, and multiclass decision tree rule generation, producing rich risk representations. A risk model is then constructed using an attention mechanism with a class-based structure designed for simultaneous multiclass risk calculation, where misprediction risk is computed jointly across all classes to capture inter-class relationships and competitive dynamics. Risk model training begins with class balancing and calibration, followed by our voting-based learning-to-rank objective that ranks images according to their risk across all classes simultaneously. This mechanism aggregates risk predictions from multiple perspectives, enabling robust multiclass risk assessment where the relative likelihood of misprediction is jointly optimized across the entire class spectrum. Finally, after identifying high-risk samples, a risk-based adaptive training approach fine-tunes a DNN model toward new workloads using temperature-scaled calibrated outputs as input. This achieves effective adaptation with minimal labeled samples by minimizing misprediction risk across both original and new distributions.


The main contributions of this work are as follows:
\begin{enumerate}
    \item We present a novel misprediction risk analysis framework for multiclass classification. Unlike prior binary-class approaches that require a separate risk model per class, our method constructs a unified, class-agnostic risk model by leveraging heterogeneous feature representations extracted from multiple deep neural network (DNN) architectures. These features are further processed and used to generate interpretable multiclass decision tree rules, enabling effective risk feature generation for the multiclass scenario.

    \item We design an attention-enhanced risk model coupled with a tailored training strategy that mitigates class bias and calibrates risk  scores for improved multiclass reliability. A voting-based learning to rank mechanism is further incorporated to enhance the effectiveness of risk model training.
    
    \item We introduce a risk-based adaptive deep learning paradigm for breast cancer subtype prediction. This approach dynamically fine-tunes the base model to a specific target workload via a two-phase process: Traditional pre-training followed by targeted, risk-based adaptive training to minimize misprediction risk.
    
    \item We perform a comprehensive empirical evaluation on multiple benchmark breast cancer histopathology datasets. Results demonstrate that the proposed risk analysis method achieves superior misprediction identification compared to existing techniques, while the adaptive learning approach consistently surpasses state-of-the-art baselines in breast cancer subtype prediction performance.
\end{enumerate}

\FloatBarrier

\section{Related Work}
In the field of breast cancer subtype prediction, recent advancements in image analysis and computer-aided diagnosis have been dominated by deep learning, especially convolutional neural networks (CNNs)~\cite{20,21,22,24,28,29,30,9,10}. Customized CNN architectures like ResHist, BreastNet, and hybrid CNN-RNN models have achieved notable accuracies, often fine-tuned on pretrained models such as VGG16 and Inception-V3. Benchmark datasets like BACH~\cite{9} and BRACS~\cite{10} have played a crucial role in standardizing evaluation metrics and facilitating the development of tailored models. While deep CNN architectures remain state-of-the-art, the field continues to evolve with the exploration of vision transformers and hierarchical graph networks~\cite{32,33}. In histopathological image-based prediction, transformer-based and multiple-instance learning (MIL) frameworks have demonstrated strong performance. TransPath~\cite{91} and CTransPath~\cite{92} leverage vision transformers to capture long-range dependencies in whole-slide images, while TransMIL~\cite{96} and CLAM~\cite{97} adopt MIL-based architectures to effectively aggregate patch-level features for slide-level classification. Furthermore, with the advancement of Large Language Models, recent works have extended vision--language pretraining to medical domains. MedCLIP~\cite{100} leverages unpaired medical image--text data for contrastive representation learning, PathCLIP~\cite{101} applies image-text contrastive learning to biomedical pathway figures, and LLaVA-Med~\cite{102} demonstrates instruction-tuned large vision--language models for biomedicine.

In the context of adaptive deep learning, various approaches have been proposed, where most of them adapt a deep model towards a target workload overcoming distribution misalignment between training and target data. Two primary strategies, transfer learning and adaptive representation learning data~\cite{39,41,42,43,45}, address this issue. Transfer learning aims to adapt a model from a source domain to perform well on a target domain, while adaptive representation learning focuses on learning domain-invariant features shared across diverse domains. Despite their effectiveness, existing approaches often struggle to fine-tune deep models for new workloads with specific characteristics. In image classification domain adaptation, recent methods like SCDA (Semantic Concentration for Domain Adaptation)~\cite{44}, TSA (Transferable Semantic Augmentation)~\cite{45}, and DANN (Domain-Adversarial Neural Network) target distribution alignment and feature augmentation, with DANN leveraging adversarial learning to reduce domain shift. These strategies can be adapted for various medical image analysis tasks. Moreover, using misprediction risk analysis in adaptive deep learning, as demonstrated in tasks like entity resolution ~\cite{18, 15} and network intrusion detection ~\cite{19}, remains a valuable direction. 

Analyzing misprediction risk, also known as "confidence ranking" or "trust scoring," has been a key focus~\cite{34,35,18,38}. Early approaches relied on softmax distributions to identify misclassified instances. More recently, the LearnRisk framework has emerged as an interpretable and trainable solution for risk analysis. Notably, our work extends LearnRisk's application from entity resolution to the domain of breast cancer subtype prediction, introducing a new design with several important strategies.

\section{Preliminaries}

\subsection{Problem Statement}
 
    The task of breast cancer subtype prediction involves assigning breast tissue images to one of several predefined categories. Given a dataset \(\mathcal{D} = \{(\mathbf{x}_i, y_i)\}_{i=1}^N\), the objective is to train a model \(M\) that accurately predicts the class label \(y \in \{1, 2, \dots, C\}\)  for feature x while minimizing prediction errors. In addition to this, misprediction risk analysis focuses on assessing the likelihood of errors in the predictive model \(M\). By generating risk features and constructing a risk model, the goal is to provide a calibrated risk score \(r\) that quantifies the probability of misprediction, enhancing model's reliability and supporting clinical decision-making.

  The purpose of this work is to create a misprediction risk analysis framework which can effectively identify the mispredictions from a breast cancer subtype prediction output. The primary goal of this work is to utilize a predictor model \(M\) to detect and predict H\&E-stained histopathological breast tissue images into multiclass categories such as: Normal, Pathological Benign, Usual Ductal Hyperplasia, Flat Epithelial Atypia, Atypical Ductal Hyperplasia, Ductal Carcinoma In Situ, and Invasive Carcinoma further grouped into three broader categories: Benign, Atypical, and Malignant. The dataset \(D\) consists of labeled training instances (\(D_s\)), labeled validation instances (\(D_v\)), and unlabeled target instances (\(D_t\)). The focus of  misprediction risk analysis is on evaluating the potential mispredictions of unlabeled target instances (\(D_t\)) across various cancer subtypes. This analysis involves ranking mispredicted pairs \((x_i^t, y_i^t)\) before correctly predicted pairs, where \(x_i^t\) represents the image and \(y_i^t\) is its associated label. The classifier \(M\) is informed about mispredictions through a risk ranking list. Evaluation metrics such as the Receiver Operating Characteristic (ROC) and the Area Under the ROC Curve (AUROC) are utilized, with the ultimate goal of maximizing the ROC metric by considering True Positive (\(TP\)), True Negative (\(TN\)), False Positive (\(FP\)), and False Negative (\(FN\)) metrics. This comprehensive approach aims to enhance the understanding of misprediction risks and improve the overall performance of the prediction model.

\subsection{The LearnRisk Framework}

The LearnRisk framework is designed to address the challenges of risk analysis in binary prediction, particularly for task of Entity Resolution. The framework operates in three key steps: \textbf{risk feature generation}, \textbf{risk model construction}, and \textbf{risk model training}. 

The first step, \textbf{risk feature generation}, involves evaluating mislabeling risks based on extracted features that satisfy three critical criteria: they must be interpretable, discriminating, and high-coverage. For instance, risk features should be easily understood by humans, clearly differentiate between true labels, and generalize well across data pairs to ensure effective transfer from training to test data. However, applying LearnRisk to multiclass problems introduces challenges, such as generating quality risk features when dissimilar images (e.g., benign vs. tumor subtypes) are grouped together or when imbalanced data complicates feature generation. Overcoming these challenges is vital for multiclass tasks, such as breast cancer prediction, where features like embedding distances might indicate risk but lack clarity in handling cases beyond specific thresholds.

Next, in \textbf{risk model construction}, the framework models equivalence probabilities for each image pair based on its risk features. Inspired by portfolio investment theory, LearnRisk treats each risk feature as a stock and each pair as a portfolio. The equivalence probability distribution for a labeled pair is represented as \( N(\mu_i, \sigma^2_i) \), where \( \mu_i \) is the expectation and \( \sigma^2_i \) is the variance. These distributions are truncated to ensure probabilities lie between 0 and 1. The framework utilizes feature weights \( w = [w_1, w_2, \dots, w_m]^T \) and feature equivalence probability distributions \( N(\mu_{f_j}, \sigma^2_{f_j}) \). The expectation and variance vectors are \( \mu_F = [\mu_{f_1}, \mu_{f_2}, \dots, \mu_{f_m}]^T \) and \( \sigma^2_F = [\sigma^2_{f_1}, \sigma^2_{f_2}, \dots, \sigma^2_{f_m}]^T \). For each pair \( d_i \), with feature vector \( x_i = [x_{i1}, x_{i2}, \dots, x_{im}] \), the distribution is defined as:

\begin{equation}
\mu_i = x_i (w \circ \mu_F),
\end{equation}

\begin{equation}
\sigma^2_i = x_i (w^2 \circ \sigma^2_F),
\end{equation}
where \( \circ \) denotes element-wise product. Value at Risk (VaR) ~\cite{46} is then employed to rank pairs based on mislabeling risks, using both expectation and variance to capture the risk of mispredictions effectively.

Finally, \textbf{risk model training} fine-tunes the parameters of the risk model to align with the characteristics of breast cancer detection. This step uses a learning-to-rank approach to rank labeled pairs by their mislabeling risk. The model parameters include feature weights (\( w_i \)), the expectation (\( \mu_i \)), and the variance (\( \sigma^2_i \)) of the risk feature distribution. The expectation is estimated from labeled data, while weights and variances are tuned during training to optimize risk ranking. The training process uses both target workload data (e.g., breast cancer detection datasets) and validation data to ensure the model captures task-specific characteristics effectively.

\section{Multi-class Risk Analysis Solution}

In this section, we present the solution for multiclass risk analysis: 

\subsection{\textbf{Risk Feature Generation}}

If we consider the binary prediction example for the breast cancer in terms of risk feature generation, we need to classify the mappings between images in labeled training data and the categories and for the binary cancer prediction the categories or classes are `Not Cancerous'' (0) and ``Cancerous'' (1), i.e. negative or positive. For example we have two images shown in table \ref{tab:true_labels_predictions}: 
\begin{table}[h!]
\centering
\begin{tabular}{ccc}
\hline
Sample & True Label & Prediction \\
\hline
1 & Not Cancerous (0) & Not Cancerous (0) \\
2 & Not Cancerous (0) & Cancerous (1) \\

\hline
\end{tabular}
\caption{True Labels and Predictions for Binary Prediction}
\label{tab:true_labels_predictions}
\end{table}

Here, Sample 1 is considered a negative sample, meaning it is not risky, while Sample 2 is considered a positive (i.e., risky) sample. The straightforward \textit{LearnRisk} solution uses a one-hot encoding scheme, mapping samples to either positive or negative values. However, this binary approach is inadequate for multiclass problems. If we directly apply the \textit{LearnRisk} solution to a multiclass task, we must mark positive and negative samples and train a separate risk model for each class.  

In the multiclass scenario, all samples with ground-truth labels different from the target label would be grouped into a single category. However, these samples may share very limited similarity, making wholesale risk analysis on them inaccurate. Moreover, the binary approach typically results in imbalanced samples, with far fewer risky samples than non-risky ones, further complicating risk analysis.  

To address this challenge in breast cancer subtype prediction (e.g., with seven subtypes), we propose marking each $\langle$sample, label$\rangle$ pair as either Match or Unmatch. As a result, each sample has one Match pair and multiple Unmatch pairs. Risk features are then generated based on these Match and Unmatch pairs. A unified risk model is subsequently constructed using these risk features to quantify the risk of machine predictions more effectively.

\begin{algorithm}[ht]
\caption{Multiclass Risk Feature Generation}
\label{alg:riskfeature}
\textbf{Input:} Dataset D and Model Architecture M \\
\textbf{Output:} Risk features for breast cancer dataset in the form of rules
\begin{algorithmic}[1]
\Procedure{FusionBasedRiskFeatureGen}{}
    \State Train models M (CNNs, Transformer) on Dataset D
    \For{each image $I$ in $D$}
        \State Extract features using the trained Deep Models.
        \State Apply feature selection (MI, FS).
        \State Fuse the features along y-axis for each label.
        \State Calculate risk metrics (CCD, KNN).
        
    \EndFor
    \State Map images to categories (M/U) based on true/predicted labels.
    \For{$i$ in range($m$)}
        \State Generate multiclass one sided decision tree rules.
        \State Generate risk features using the mappings and rules with fused features.
    \EndFor
\EndProcedure
\end{algorithmic}
\end{algorithm}

To tackle challenges in breast cancer subtype prediction, we developed a solution for risk feature generation that optimizes multi-class model performance while ensuring high coverage and interpretability.  We have sketched the full workflow of risk feature generation in Algorithm \ref{alg:riskfeature}, whose technical details will be described next. 


In our solution, we utilize multiple DNN models to extract diverse features, while enhancing representation, that can also lead to data ambiguity and overfitting, especially in multiclass scenarios where features are often correlated and redundant. To mitigate these challenges, we implement two correlation-based feature selection techniques: Mutual Information~\cite{52} and F-Score~\cite{53}. Mutual Information quantifies the dependence between features \(X\) and the target variable \(Y\) as defined by Equation~\ref{eq:mutual_information}:

\begin{equation}
\label{eq:mutual_information}
I(X; Y) = \sum_{x \in X} \sum_{y \in Y} p(x,y) \log\left(\frac{p(x,y)}{p(x)p(y)}\right)
\end{equation}
where \(p(x, y)\) is the joint probability distribution of \(X\) and \(Y\), and \(p(x)\) and \(p(y)\) are the marginal probability distributions of \(X\) and \(Y\), respectively.
The F-score assesses feature discriminative power by comparing variance between and within classes, as shown in Equation~\ref{eq:f_score}:

\begin{equation}
\label{eq:f_score}
F(f_i) = \frac{\sum_{j} (\mu_i - \mu_{i,j})^2}{\sum_{j} \sigma_{i,j}^2}
\end{equation}
where \(\mu_i\) is the overall mean of feature \(f_i\), \(\mu_{i,j}\) is the mean of feature \(f_i\) within class \(j\), and \(\sigma_{i,j}^2\) is the variance of feature \(f_i\) within class \(j\). For example, in feature selection from ResNet-50, let \(\mathbf{f}_{\text{ResNet-50}} \in \mathbb{R}^{d}\) be the output feature vector. We define selection functions:

\begin{equation}
\textstyle
F_{\text{MI}}(\mathbf{f}_{\text{ResNet-50}}) = \text{Top}_{200}(I(\mathbf{f}_{\text{ResNet-50}}; Y))
\end{equation}

\begin{equation}
\textstyle
F_{\text{FS}}(\mathbf{f}_{\text{ResNet-50}}) = \text{Top}_{200}(F(\mathbf{f}_{\text{ResNet-50}}))
\end{equation}
where \(I(\mathbf{f}_{\text{ResNet-50}}; Y)\) is the Mutual Information between the feature vector and target labels, \(F(\mathbf{f}_{\text{ResNet-50}})\) represents the F-score of the feature vector, and \(\text{Top}_{200}\) selects the top 200 features based on their respective scores.

Finally, we perform feature fusion via concatenation of selected features from multiple models and both feature selections, resulting in:

\begin{equation}
\mathbf{F}_{\text{Fused}} = [\mathbf{f}_{M_1}, \mathbf{f}_{M_2}, \ldots, \mathbf{f}_{M_i}], 
\label{eq:feature_fusion}
\end{equation}
where $M_i$ denotes a DNN model and $f_{M_i}$denotes the selected features of $M_i$.

   Based on the results of feature fusion on an image’s vector representation we extract two risk metrics of category similarity: 
   \begin{itemize}	
    \item \textbf{Category Cosine Distance (CCD):} We estimate the class center of each category by its representatives in the training data. For a given image, we measure its similarity with each category by calculating the distance to the class centroid, defined as \( 1.0 - \) vector cosine similarity. A smaller distance indicates a higher probability that the image belongs to the category. This metric is denoted as \emph{CCD} shown in equation ~\ref{eq:ccd}: 
    \begin{equation}
    \label{eq:ccd}
    \text{CCD} = 1 - \cos\left(\theta\right), \quad \text{where} \quad \cos(\theta) = \frac{\mathbf{v} \cdot \mathbf{c}}{\|\mathbf{v}\| \|\mathbf{c}\|}
    \end{equation}
    
	where v is the query vector, c is the class centroid vector, and \(\cos(\theta)\) is the cosine similarity between them.
	\item \textbf{K-nearest Neighborhood (KNN):} For each image, we identify the categories of its k-nearest neighbors in an embedding space, where nearness is determined by vector cosine similarity. The effectiveness of this metric relies on the assumption that images appearing close in an embedding space likely share the same label. In our approach, we use different values of \( K \) (e.g., \( K = 5 \) or \( K = 7 \)) to derive distinct risk metrics. With \( K = 5 \), we consider only the nearest neighbor; with \( K = 7 \), we include a broader set of representative samples from the local neighborhood. This metric is denoted as \emph{KNN} as shown in equation ~\ref{eq:knn}:  
    \begin{equation}
    \label{eq:knn}
    \mathcal{K}_{\text{NN}} = \{ \mathbf{c}_i \,|\, i \in \text{argmin}_{j \in \{1, 2, \dots, N\}} \|\mathbf{v} - \mathbf{v}_j\|, \, j = 1, \dots, k \},
    \end{equation}
    
    where, $v$ is the query       vector, $v_j$ is the vector in the embedding space, $c_i$ is the corresponding category, and k is the number of nearest neighbors.
\end{itemize}


To generate unified risk features, we construct the mappings between images in labeled training data and their corresponding categories (classes) by categorizing them as Matching (M) or Unmatching (U), where for each image there is one M and all other are U. An illustrative mapping table has been shown in Table \ref{tab:mappings}.
Leveraging this mapping information. One-sided decision trees will then be employed to extract risk features (rules) based on these mappings. The rules are generated using thresholds on risk metrics, where each partitioning focuses on maximizing the "purity" of either the Matching (M) or Unmatching (U) sets. 


This process mirrors the binary one sided decision tree rule generation, but it is adapted for multiclass settings by ensuring that each image's classification is evaluated against all possible class labels. The technical details for decision tree rules of this process align with prior work~\cite{18}. These rules operate on conditions and thresholds involving the risk metrics. It should be mentioned here that M/U targets the predicted class of an image. An illustrative example \ref{ex:onesided_tree_rule} of risk feature is shown as follows:

\begin{table}
    \centering
    \caption{Example mappings between images and categories. FF represents the Fused Features, CCD category cosine risk metric, KNN(5) denotes the KNN risk metric with \(K=5\), while KNN(7) denotes the KNN risk metric with \(K=7\).}
    \scriptsize
    \setlength{\tabcolsep}{1.5pt} 
    \label{tab:mappings}
    \begin{tabular}{lccccccc}
        \toprule
        Image & True Label & Category & M/U & FF-CCD & FF-KNN(5) & FF-KNN(7) \\
        \midrule
        $d_1$ & 0 & Positive & M & 0.06 & 5 & 7 \\
        & 0 & Negative & U & 0.53 & 0 & 0 \\
        $d_2$ & 1 & Positive & U & 0.57 & 1 & 2 \\
        & 1 & Negative & M & 0.04 & 4 & 5 \\
        $d_3$ & 2 & Positive & M & 0.07 & 4 & 6 \\
        & 2 & Negative & U & 0.63 & 1 & 1 \\
        $d_4$ & 3 & Positive & U & 0.80 & 2 & 3 \\
        & 3 & Negative & M & 0.02 & 3 & 4 \\
        \bottomrule
    \end{tabular}
\end{table}

\begin{example}
\label{ex:onesided_tree_rule}
Suppose a one sided decision tree rule based risk feature is expressed as:
\begin{equation}
d_1 [FF-CCD] < 0.11 \rightarrow M  \label{eq:riskfeature}  
\end{equation}
here, $d_1$ represents an image, CCD is the category cosine distance metric, and $d_1 [FF-CCD]$ denotes the image's cosine distance to a class centroid based on the Fused features. This rule implies that if an image, according to the Feature FF, is very close to a class center but has a machine label different from the class, it is considered at high risk of misprediction. However, it's essential to note that this rule does not indicate any category preference if the measured distance exceeds 0.11. Valid risk features, represented as rules, are statistically satisfied by a high percentage (e.g., 95\%) of instances in the training data mapping table.

\end{example}

More examples of one-sided rules generated on the BRACS benchmark dataset~\cite{10} are presented in Table~\ref{tab:rules}, in which some of the examples have also been described in example \ref{ex:tree_rules}:

\begin{table}
    \centering
    \caption{Examples of generated risk features on the BRACS Breast Cancer Dataset.}
    \label{tab:riskfeatures}
    \footnotesize 
    \setlength{\tabcolsep}{3pt} 
    \label{tab:rules}
    \begin{tabular}{lll}
        \toprule
        Rule & Description \\
        \midrule
        $r_1$: FF - 0 - CCD $\leq$ 0.104 $\rightarrow$ M & $r_2$: FF- 1 - CCD $>$ 0.359 $\rightarrow$ U \\
        $r_3$: FF - 0 - KNN5 $>$ 4.5 $\rightarrow$ M & $r_4$: FF - 2 - KNN5 $\leq$ 0.5 $\rightarrow$ U \\
        $r_5$: FF - 0 - KNN7 $>$ 6.5 $\rightarrow$ M & $r_6$: FF - 3 - KNN7 $\leq$ 0.5 $\rightarrow$ U \\
        \bottomrule
    \end{tabular}
\end{table}

\begin{example}
\label{ex:tree_rules}
Let’s take $r_1$, $r_2$, and $r_5$ as examples. $r_1$ shows if the CCD distance for the FF less than or equal to 0.104, then it is a match (M) w.r.t whatever predicted label of an image. $r_2$ shows that if the CCD distance for the FF feature is greater than 0.359, then it’s an unmatch (U), which means in this scenario, it doesn’t belong to a predicted label.
\end{example}


\subsection{\textbf{Attention-based Risk Model Construction}}

The existing binary model supposes risk features are mutually independent.  However, in real scenarios, risk features are to a large extent correlated, which is especially true in the multi-class scenario. Therefore, we present a new dynamic risk model with attention-based weighting for multi-class risk analysis.  The attention mechanism dynamically adjusts the importance of risk features based on the context of each class, improving multiclass prediction by weighting features according to their relevance~\cite{77,78}. 


Unlike the static methods of Gaussian weighting proposed for binary analysis, the proposed risk model assigns context-specific weights, enabling the model to emphasize informative data and reduce irrelevant information.  
\begin{figure}
    \centering
    \includegraphics[width=1.0\columnwidth]{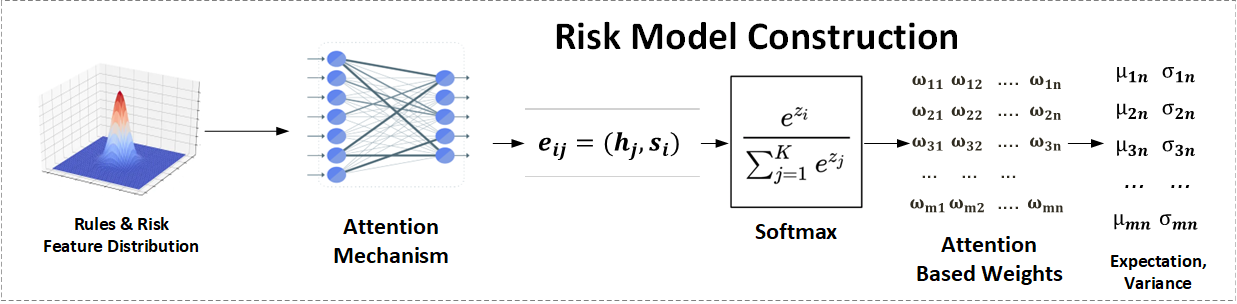}
    \caption{Attention Based Risk Model}
    \label{fig:attention}
\end{figure}
\FloatBarrier
The structure of the attention-based risk model has been presented in Figure \ref{fig:attention}. To implement the attention mechanism, we add a linear layer to the model to enable dynamic and context-sensitive feature weighting.  The linear layer learns to transform the input features \( \mathbf{x} \in \mathbb{R}^{\text{input\_dim}} \) into class-specific scores. These scores are represented as:
\begin{equation}
    e_{ij} = \text{score}(h_i, s_j),
\end{equation}
where \( h_i \) is the hidden state of the input sequence at position \( i \) and \( s_j \) is the hidden state of the output sequence at position \( j \). We then align these scores using the softmax function:
\begin{equation}
    \omega_{ij} = \frac{\exp(e_{ij})}{\sum_{k=1}^n \exp(e_{ik})}
\end{equation}
where \( \omega_{ij} \) represents the alignment score, obtained using the softmax function to normalize the scores across all input positions, which gives us the attention-based weights. Finally, we obtain context-aware expectation and variance vectors. Accordingly, the distribution of an image \( d_i \) is estimated by the context-aware vectors:
\begin{equation}
\mu_i = \mathbf{\omega} \circ \boldsymbol{\mu}_{F}
\end{equation}
\begin{equation}
\sigma_i^2 = \mathbf{\omega}^2 \circ \boldsymbol{\sigma}^2_{F}
\end{equation}
where $\omega$ represents the attention weights, $\mu$ and $\sigma^2$ represents the expectation and variance, and \( \circ \) represents the element-wise product. Different classes may require attention to different features or parts of the input. An attention mechanism can learn to focus on the most informative features for each class dynamically, improving the model’s ability to differentiate between classes.

\begin{example} 
In a multi-class prediction task, consider a sample \( d_i \) that can belong to one of four classes: A, B, C, or D. In the learnable Gaussian model, each class has a set of mean (\( \mu_A, \mu_B, \mu_C, \mu_D \)) and variance (\( \sigma_A^2, \sigma_B^2, \sigma_C^2, \sigma_D^2 \)) parameters that are learned from the risk features. These parameters are used to compute the Gaussian weights, however, even with these learnable parameters, the model still uses fixed, class-specific parameters that do not change dynamically based on the sample content. In contrast, the attention mechanism assigns dynamic weights based on the risk features. For example, for image \( d_i \), the attention mechanism might assign \( \omega_A = 0.6 \), \( \omega_B = 0.1 \), \( \omega_C = 0.2 \), and \( \omega_D = 0.1 \), emphasizing class A more than the others based on the sample's specific features. This dynamic adjustment allows the attention mechanism to focus on the most relevant classes for each input sample, whereas the Gaussian model is constrained by pre-learned, static parameters. 


\end{example}

\subsection{\textbf{Risk Model Training}}

For risk model training, our proposed approach considers the expectation (\(\mu_i\)) as prior knowledge, which can be estimated based on the labeled training data, and fine-tunes  the variances ($\sigma_{i,c}^2$) and feature weights ($\omega_i$) to differentiate between correctly and incorrectly labeled samples. We have sketched the algorithm for risk model training in Alg~\ref{alg:multiclass_risk_adjustment}.


Class imbalance is common in most histopathology image datasets, as shown in Tables \ref{tab:sevenclassdataset} and \ref{tab:fourclassdataset}, particularly in the training data. To address this issue, we propose to neutralize class bias in expectation and variance for balanced learning and prediction. The neutralization operation is formally defined by Equation \ref{eq:adjusted_mean}, and \ref{eq:adjusted_variance} as follows:

\begin{equation}
\mu_i = softmax(\mu_{\text{raw},i} - w_i)
\label{eq:adjusted_mean}
\end{equation}

\begin{equation}
\sigma_i^2 = \sigma_{\text{raw},i}^2 - \alpha \cdot w_i
\label{eq:adjusted_variance}
\end{equation}
where $\mu_{\text{raw},i}$ and $\sigma_{\text{raw},i}^2$ represent the raw expectation and variance values, $w_i$ the class weights, and $\alpha$ the scaling factor. The adjusted expectations $\mu_i$ are then normalized using the softmax function. 

 

Furthermore, compared to binary risk analysis, where there are only two candidate classes,
the multi-class scenario usually has less labeled data for risk model training for each class. To address ineffectiveness of labeled training data and class imbalance,  we use Platt Scaling calibration to adjust classifier overconfidence ~\cite{84} as follows:

\begin{equation}
P(y = k \mid x) = \frac{1}{1 + \exp(A_k \cdot f_k(x) + B_k)}
\end{equation}
where \(P(y = k \mid x)\) is the probability of class \(k\) given input \(x\), \(f_k(x)\) is the output score of the classifier for class \(k\), and \(A_k, B_k\) are calibration parameters learned during the Platt Scaling process. Calibration enhances decision accuracy by aligning predicted risk with actual outcomes, using class-specific parameters \( A_k \) and \( B_k \) along with the raw score \( f_k(x) \).
\begin{algorithm}[!ht]
\caption{Multiclass Risk Model Training }
\label{alg:multiclass_risk_adjustment}
\begin{algorithmic}[1]
\Require
\State Positive Samples
\State Pairs of data and their respective classes: $(d_i, c_i), (d_j, c_j)$
\State Initial expectations and variances for each class: $\mu_{i,c}$, $\sigma_{i,c}^2$
\State VaR risk values for each class: $\text{VaR}_c$
\State Learning rate $\eta$
\State Maximum number of training iterations $N$
\Ensure 
\State Updated expectations $\mu_{i,c}$, variances $\sigma_{i,c}^2$, attention weights $\omega$, class weights $w$, and Risk Calibration

\For {$i = 1$ to $N$}
    \State Calculate Attention-based Weights $\omega_{ij}$
    
    \State Update Context-aware Expectation and Variance: $\mu_{i,c} \gets \mathbf{\omega} \circ \boldsymbol{\mu}_F(c)$, $\sigma_{i,c}^2 \gets \mathbf{\omega}^2 \circ \boldsymbol{\sigma}^2_F(c)$

     \State Neutralize Class Bias: $\mathbf{\mu}_i = \mathbf{\mu}_{\text{raw}, i} - w_{i}$, 
     $\sigma_i^2 = \sigma_{\text{raw},i}^2 - \alpha \cdot w_i$
     \State Risk Calibration using platt scaling:
        $1 + \exp(A_k \cdot f_k(x) + B_k)$   
    \State \textbf{Learning to Rank:}
        \State Calculate the desired target probability $\overline{p}_{ij, c}$ 
        \State Calculate the posterior probability $p_{ij, c}$ 
        
        \State Voting based Ranking:
            \[
            W_i = \sum_{d_i, c_i} \text{Indicator}(\gamma_{i, c_i} > \gamma_{j, c_j})
            \]
        
        \State Calculate the loss using aggregated scores $L(D_\gamma)$

\EndFor

\State Calculate the final loss $L(D_\gamma)$
\State \textbf{return} Risk Ranking List
\end{algorithmic}
\end{algorithm}


In multiclass risk analysis, given two images, $d_i$ and $d_j$, and the mislabeling risk of  $(d_i, c_i)$ and $(d_j, c_j)$, is $\gamma_{i, c_i}$ and $\gamma_{j, c_j}$., the ranking notation can be denoted as:

\begin{equation}
(d_i, c_i) \triangleleft (d_j, c_j) \quad \text{if and only if} \quad \gamma_{i, c_i} > \gamma_{j, c_j}
\end{equation}
where $\gamma$ denotes the risk score of the particular classes of corresponding samples. Considering this scenario, the posterior probability $p_{ij, c}$ of the image $(d_i, c_i)$ is ranked higher than $(d_j, c_j)$ is given by:

\begin{equation}
\label{eq:post-probability}
p_{ij, c} = \frac{e^{\gamma_{i, c_i} - \gamma_{j, c_j}}}{1 + e^{\gamma_{i, c_i} - \gamma_{j, c_j}}}
\end{equation}
and the desired target value for the posterior probability is now given by:

\begin{equation}
\label{eq:target_probability}
\overline{p}_{ij, c} = 0.5 \times (1 + \hat{g}_{i, c_i} - \hat{g}_{j, c_j}),
\end{equation}
where $\hat{g}_{i, c_i} $ and $\hat{g}_{j, c_j} $$\in \{0, 1\}$ denote the risk labels of pairs. If a pair $\hat{g}_{i, c_i} $ is mislabeled, $\hat{g}_{i, c_i} = 1$; otherwise, $\hat{g}_{i, c_i}  = 0$.
The objective is to minimize the difference between $p_{ij}$ and $\overline{p}_{ij}$. The straightforward way to optimize the objective is to perform comparisons over all the images with various classes. However, this approach may be biased if certain classes have higher risk score and it does not account for the relative performance of each class. 

     
     Therefore, rather than using the individual-based approach, we present a voting-based mechanism for ranking images. For each item \((d_i, c_i)\), we count the number of wins, defined as the instances where \((d_i, c_i)\) is ranked higher than \((d_j, c_j)\):

\begin{equation}
W_i = \sum_{d_i, c_i} \text{Indicator}(\gamma_{i, c_i} > \gamma_{j, c_j})
\end{equation}
here, \(\text{Indicator}(\cdot)\) returns 1 if the condition is met and 0 otherwise. In case of a tie between class votes, our method resolves the ranking by prioritizing the pair with the highest aggregated risk score. This approach results in class-specific and balanced rankings for each image, prioritizing those with a higher number of risky classes. The example below further describes the process:

\begin{example}
In a breast cancer subtype prediction task with four classes (benign, malignant, in situ, and invasive), consider two images, \(d_i\) and \(d_j\), with predicted class probability vectors \(d_i = [0.2, 0.6, 0.1, 0.1]\) and \(d_j = [0.3, 0.4, 0.2, 0.1]\). Performing pairwise comparisons, \(d_i\) wins against \(d_j\) in the malignant class (1 win) but loses in the benign and in situ classes (0 wins each), while both images tie in the invasive class. The total win count results in \(W_i = 1\) for \(d_i\) and \(W_j = 2\) for \(d_j\). Consequently, \(d_j\) is ranked higher due to its superior win count, demonstrating the effectiveness of the voting-based mechanism for class-based image ranking.

\end{example}

If $(d_i, c_i)$ is mislabeled and $(d_j, c_j)$ is correctly labeled, the optimization process will maximize $\gamma_{i, c_i} - \gamma_{j, c_j}$ for all the class using the aggregated $p_{ij}$ and $\overline{p}_{ij}$ scores in a multiclass scenario, the cross-entropy loss function is extended to account for multi-class pairs as follows:

\begin{equation}
\label{eq:loss_function}
L(D_\gamma) = \sum_{(d_i, c_i), (d_j, c_j) \in D_\gamma} -\overline{p}_{ij} \log(p_{ij}) - (1 - \overline{p}_{ij}) \log(1 - p_{ij}),
\end{equation}
in which the value of \(L(D_{\gamma})\) increases with the value difference between \(p_{ij}\) and \(\bar{p}_{ij}\) as desired.

We also introduced dropout during training to randomly set a percentage of input units to zero. This prevents co-adaptation and promotes the learning of more robust features for each class. Dropout is inactive during evaluation, with activations scaled for consistency. In multiclass prediction, dropout helps prevent overfitting by ensuring that the model does not overly rely on specific features. We set the dropout rate to 0.5 and fine-tuned it for optimal performance.


\section{Solution for MultiRisk Based Adaptive Training}

For the task of breast cancer subtype classifcation, neural networks using softmax often produce overly confident predictions, which can mislead training and impair generalization~\cite{73}. Class imbalance can further degrade performance on minority classes~\cite{74}, and risk estimation may fail to accurately reflect misprediction likelihood, particularly in complex multiclass scenarios~\cite{35}.

Our solution, detailed in Algorithm \ref{alg:risk_adaptive_bcd} and Figure \ref{fig:risk_analysis}, involves two phases: traditional pre-training and risk-based adaptive training.

\begin{itemize}
    \item \textbf{Traditional Pre-Training Phase}: During this initial phase, a deep model undergoes training using conventional methods, relying on labeled training data for breast cancer subtypes. In our method, we use four CNN deep models and train all of them on the given workload with arguably the best parameters. The aim is to try to achieve the best test efficacy for each model.
    \item \textbf{Risk Based Adaptive Training Phase}: This phase aims to enhance model performance by focusing on high-risk instances and managing prediction uncertainties. This framework integrates temperature scaling to reduce overconfidence and Value at Risk (VaR) to assess misprediction risks. The goal is to improve the robustness and accuracy of the prediction model. The step-wise process is given below:
\end{itemize}

\begin{algorithm}[!ht]
\caption{Risk-Based Adaptive Training for Breast Cancer Subtype Prediction}
\label{alg:risk_adaptive_bcd}
\textbf{Input:} A workload of breast cancer subtype histopathological images \(D\) consisting of \(D_s\), \(D_v\), \(D_t\) and initial classifier \(g(\omega)\); \\
\textbf{Output:} A learned classifier \(g(\omega_*)\);
\begin{algorithmic}[1]
\State \(\omega_0 \leftarrow\) Initialize \(\omega\) with random values;
\For{\(k=0\) to \(m-1\)}
    \State Train the classifier using the standard cross-entropy loss on the training set \(D_s\):
    \State \(\omega_{(k+1)} \leftarrow \omega_{(k)} - \alpha \nabla_{\omega_k} L_{\text{train}}(\omega_k)\);
\EndFor
\State Select the best model \(g(\omega_*)\) based on validation set \(D_v\);
\State \(\omega_m \leftarrow \omega_*\);
\State Train the risk model using the validation set \(D_v\):
    \For{each instance \(i \in D_v\)}
        \State Calculate \(\text{VaR}_{\theta}(i)\) for each class;
    \EndFor
\For{\(k=m\) to \(m+n-1\)}
    \State Update the model using the test set \(D_t\) with labels predicted by the risk model:
    \For{each instance \(i \in D_t\)}
        \State Predict the probabilities \(P_{\text{original}}\) for each class;
        \State Regularize the predictions:
        \State \( P_{\text{regularized}}(i) = \frac{\exp(z_i / \lambda)}{\sum_{j=1}^C \exp(z_j / \lambda)} \);
        \State Compute the cross-entropy loss \(L\) using \(P_{\text{regularized}}\) and true labels;
    \EndFor
    \State Update the model parameters:
    \State \(\omega_{(k+1)} \leftarrow \omega_{(k)} - \alpha \nabla_{\omega_k} L(\omega_k)\);
    \State Recalculate the risk values and update \(\lambda\);
\EndFor
\State Select the last trained model \(g(\omega_*) \leftarrow g(\omega_{(k+1)})\);
\State \textbf{return} \(g(\omega_*)\);
\end{algorithmic}
\end{algorithm}

In the second phase, the algorithm first uses the validation set \(D_v\) to train the risk model, focusing on estimating the Value at Risk (VaR) for each instance to aid in risk assessment. This involves identifying high-risk instances through VaR calculations for additional training or correction. Then, it applies the trained risk model to predict labels for the test set \(D_{test}\), assigning tentative labels based on the lowest risk as indicated by VaR. Finally, it fine-tunes the classifier using the instances in \(D_{test}\) and their predicted labels, with the goal of minimizing misprediction risk as guided by VaR.


Observing that models relying on softmax outputs often produce overconfident predictions, we propose to smooth the predicted probability distribution for improved calibration. Guo et al.~\cite{75} highlighted that techniques like temperature scaling can enhance calibration without complex calculations. By applying logits scaling with a custom softmax activation function that incorporates a learnable temperature parameter \(\lambda\), we adaptively scale the logits, as shown in equation \ref{eq:regularized_softmax}, as follows:

\begin{equation}
    P_{\text{regularized}}(i) = \frac{\exp(z_i / \lambda)}{\sum_{j=1}^C \exp(z_j / \lambda)},
    \label{eq:regularized_softmax}
\end{equation}
where the parameter \(\lambda\) is initialized at 2 and updated during training to balance regularization. Here, \(z_i\) represents the logits for class \(i\), \(C\) is the total number of classes, and \(\exp\) denotes the exponential function. We use cross-entropy loss to calculate the difference between predictions and true targets, while the SGD optimizer adjusts \(\lambda\) based on gradients from back-propagation. Iterative updates in the training loop minimize the loss, ensuring effective learning, the process is represented as:

\begin{equation}
    L(Y, P_{\text{regularized}}, \lambda) = -\frac{1}{N} \sum_{i=1}^{N} Y_i \log(P_{\text{regularized}}(i))
    \label{eq:regularized_loss}
\end{equation}
where \(L\) is the Cross Entropy Loss applied to the adjusted probabilities \(P_{\text{regularized}}\). \(Y_i\) is the one-hot encoded true label for the \(i\)-th sample, \(N\) is the total number of samples in the batch, and \(\log\) is the natural logarithm. The parameter \(\lambda\) regulates the temperature of the softmax probabilities, improving regularization, and is iteratively updated during training using gradients.

\section{Empirical Study}

This section evaluates the performance of our proposed approach by an empirical study. It is organized as follows: Subsection.~\ref{sec:setup} describes experimental setup; Subsection.~\ref{sec:RiskAnalysis} presents the evaluation results of risk analysis; Subsection.~\ref{sec:AdaptiveLearning} presents the evaluation results of adaptive learning; Subsection.~\ref{sec:Sensitivity} presents the evaluation results of sensitivity analysis;
Subsection.~\ref{sec:ablation} presents the evaluation results of ablation study;
Subsection.~\ref{sec:generalization} presents the evaluation results of generalization on other type of cancers;
and finally, Subsection.~\ref{sec:scalability} presents the presents the discussion on computational efficiency and scalability of our method;


\subsection{Experimental Setup} \label{sec:setup}


\begin{table}[ht]
    \centering
    \caption{The statistics of the BRACS dataset}
    \label{tab:sevenclassdataset}
    \setlength{\tabcolsep}{3.0pt} 
    \begin{tabular}{lcccccccc}
        \hline
        \textbf{}       & \textbf{N} & \textbf{PB} & \textbf{UDH} & \textbf{FPA} & \textbf{ADH} & \textbf{DCIS} & \textbf{IC}   & \textbf{Total} \\
        \hline
        \textbf{Train}  & 357        & 714         & 389          & 624          & 387          & 665           & 521           & 3657           \\
        \textbf{Validation} & 46         & 43          & 46           & 49           & 41           & 40            & 47            & 312            \\
        \textbf{Test}       & 81         & 79          & 82           & 83           & 79           & 85            & 81            & 570            \\
        \hline
    \end{tabular}
\end{table}

\begin{table}[ht]
    \centering
    \caption{The statistics of the BACH Dataset}
    \label{tab:fourclassdataset}
    \begin{tabular}{lccccc}
        \hline
        \textbf{Dataset} & \textbf{Normal} & \textbf{Benign} & \textbf{DCIS} & \textbf{IC} & \textbf{Total} \\
        \hline
        \textbf{Train}    & 357             & 650             & 665          & 521         & 2193          \\
        \textbf{Validation} & 16              & 16              & 16           & 16          & 64            \\
        \textbf{Test}       & 20              & 20              & 20           & 20          & 80            \\
        \hline
    \end{tabular}
\end{table}

\begin{figure*}
    \centering
    \includegraphics[width=1.0\textwidth]{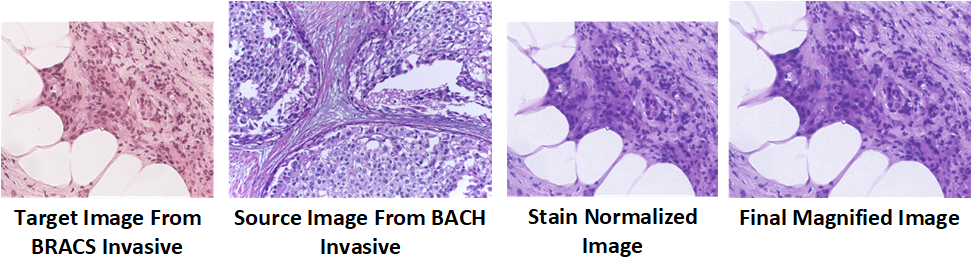}
    \caption{Stain Normalization on BRACS}
    \label{fig:stain_normalization}
\end{figure*}

\begin{figure}
    \centering
    \includegraphics[width=0.6\linewidth]{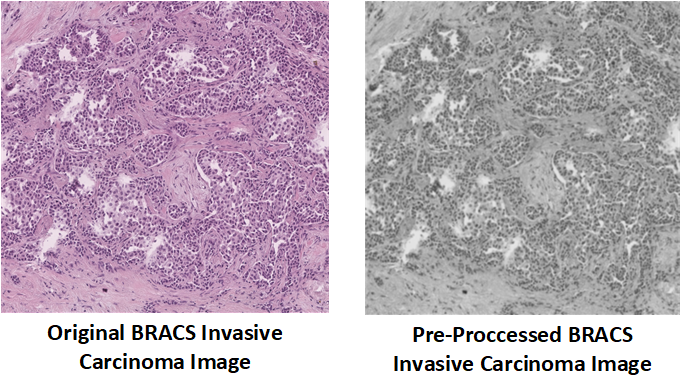}
    \caption{Sample for Original image and Low-Resolution Image from Invasive Carcinoma Class of BRACS}
    \label{fig:sample_images}
\end{figure}
We have used two open-sourced benchmark datasets of real H\&E stained histopathological breast cancer images, BRACS~\cite{10} and BACH~\cite{9}. They can be briefly described as follows:  

\begin{itemize}
    \item \textbf{BRACS.} The BReAst Carcinoma Sub-typing (BRACS) dataset is a large, annotated collection of 547 Whole-Slide Images (WSIs) and 4539 Regions Of Interest (ROIs) from over 150 patients, aimed at advancing AI for breast cancer characterization. Annotated by three pathologists, the dataset categorizes lesions into three categories and seven subtypes, making it the most comprehensive dataset available for breast cancer sub-typing. Table~\ref{tab:sevenclassdataset} shows the statistics of this dataset.
    \item \textbf{BACH.} The BACH dataset from the ICIAR 2018 grand challenge contains 400 ROI microscopy images, 100 for each of its four classes. Due to its limited size, we combined BACH with the BRACS dataset, as both share similar image types. For adaptive learning evaluation, we used BRACS for training and BACH for validation and testing. Table~\ref{tab:fourclassdataset} shows the statistics of this dataset.
\end{itemize}

Additionally, we have converted the original BRACS dataset to a low-resolution version, creating a third dataset, as CNNs and Transformers perform better on low-resolution images. Our experiments have been conducted on the original BRACS with varying image sizes, low-resolution BRACS, and the BRACS-BACH datasets.

For the evaluation of cross-domain risk analysis and adaptive learning, we used the BRACS and BACH~\cite{9} breast histopathological datasets. We combined them into a new four-class dataset (Normal, Benign, In Situ, and Invasive), merging BRACS's benign and hyperplasia classes into one. Macenko stain normalization~\cite{59} addressed staining differences, and BRACS images were resized to 20x magnification for uniformity. The dataset was balanced by selecting proportionate images from each class. For the seven-class task, we used both the original and a low-resolution grayscale BRACS dataset at 512 x 512 pixels. Figure \ref{fig:stain_normalization} and \ref{fig:sample_images} shows a sample of preproccessing on given images.

Furthermore to demonstrate the generalizability of our approach beyond breast cancer, we also conducted experiments on two additional cancer histopathology datasets: LC25000~\cite{98}, which contains 25,000 images of lung and colon cancer across five classes including lung adenocarcinoma, lung squamous cell carcinoma, colon adenocarcinoma, lung benign, and colon benign tissues; and LungHist700~\cite{99}, comprising 700 images with three classes focusing on lung adenocarcinoma, lung squamous cell carcinoma, and normal lung tissue.

For misprediction risk analysis, we train several state-of-the-art models, including Densenet121, Densenet201, Resnet50, EfficientnetB4, and the transformer-based CCT; then, we choose the top-performing model as the classifier baseline, while using the others to generate diverse risk features. All the models were trained for 200 epochs while for Resnet50, Densenet121, Densenet201 and CCT, the initial learning rate was set to 0.0005 and for EfficientnetB4, the learning rate was set to 0.001. Furthermore, SGD optimizer and Cross Entropy loss function were employed to optimize the performance of the models. The Learning rate for risk model training was set between the range from 0.0005 to 0.00005.



    For misprediction risk analysis, we compare the proposed approach, denoted by MultiRisk with the following SOTA approaches: 

\begin{itemize}
  \item \textbf{Baseline}~\cite{34}: This technique evaluates uncertainty by focusing on the classifier's output probability. Predictions near 1 indicate confidence, while those closer to 0 imply uncertainty. The uncertainty score is calculated as \(1 - P(e_{ij} | m_i)\), emphasizing uncertainty over confidence.

  \item \textbf{TrustScore}~\cite{35}: TrustScore calculates a score based on the agreement between the classifier and a nearest-neighbor classifier using features from DCA’s layer. We use the official implementation with default settings.

  \item \textbf{ConfidNet}~\cite{60}:ConfidNet adds a confidence scoring module to the main network. This 5-layer module with ReLU activations computes the confidence score, trained using the True Class Probability (TCP) criterion.

  \item \textbf{LearnRisk} ~\cite{18}: This approach implements the straightforward LearnRisk solution for multi-class classification, constructing a separate risk model for each class. 

\end{itemize}

For the evaluation of risk-based adaptive learning, we compare our solution with state-of-the-art methods from the perspective of multiple categories, inlcuding backbones, SOTA multiclass breast cancer and histopathological image classification, domain adaptation, and uncertainty estimation-based training:

\begin{itemize}
    \item \textbf{Backbones:}
    \begin{itemize}
        \item \textbf{CNNs:} Multiple CNN architectures were used to extract features for the MultiRisk approach.
        \item \textbf{CCT~\cite{63}:} The Compact Convolutional Transformer merges the strengths of CNNs and transformers, achieving top performance in tasks such as image classification, object detection, and semantic segmentation.
    \end{itemize}

    \item \textbf{SOTA Breast Cancer \& Histopathological Image Prediction:}
    \begin{itemize}
        \item \textbf{HactNet~\cite{61}:} Creates Hierarchical Cell-to-Tissue (HACT) graph representations for predicting subtypes from ROI images and compares results with pathologists' predictions.
        \item \textbf{ScoreNet~\cite{32}:} Transformer-based method using scoremix augmentation to achieve top prediction performance on BRACS ROIs, outperforming techniques like multiple instance learning and GNNs.
        \item \textbf{TransPath~\cite{91}:} Transformer-based model utilizing attention mechanisms to effectively capture complex patterns and relationships in histopathological image data.
        \item \textbf{CTransPath~\cite{92}:} Employs Semantically-Relevant Contrastive Learning (SRCL) in a hybrid CNN–Swin Transformer model to improve histopathological image tasks.
        \item \textbf{TransMIL~\cite{96}:} A transformer-based Multiple Instance Learning framework tailored for whole-slide image classification in pathology.
      \item \textbf{CLAM~\cite{97}:} Clustering-constrained Attention Multiple Instance Learning framework that provides interpretable histopathology classification.
    \end{itemize}

    \item \textbf{Domain Adaptation:}
    \begin{itemize}
        \item \textbf{SCDA~\cite{44}:} Semantic Concentration for Domain Adaptation focuses on key features through pair-wise adversarial alignment of prediction distributions.
        \item \textbf{TSA~\cite{45}:} Transferable Semantic Augmentation improves classifier adaptation by augmenting source features with semantically-relevant random directions.
        \item \textbf{DANN~\cite{95}:} Domain-Adversarial Neural Network using gradient reversal to learn domain-invariant representations for improved generalization.
    \end{itemize}

    \item \textbf{Uncertainty Estimation(Risk)–Based Training:}
    \begin{itemize}
        \item \textbf{MCdropout~\cite{93}:} Monte Carlo dropout is applied during inference to estimate predictive uncertainty by sampling multiple stochastic forward passes.
        \item \textbf{Deep Ensemble~\cite{94}:} Combines predictions from multiple independently trained models to capture both model and data uncertainty for more robust predictions. In our case three CNNs combined with CCT
        \item \textbf{LearnRisk~\cite{18}:} Uses a Risk Model for misprediction risk analysis and adaptive training to enhance the performance of state-of-the-art approaches.
    \end{itemize}
    \item \textbf{MultiRisk(Ours):} Uses misprediction risk analysis designed for subtype prediction and adaptive training to enhance the performance of state-of-the-art approaches.
\end{itemize}

   Furthermore, we conducted experiments using large language model–based vision-language models, using the concept of CLIP~\cite{104}, and DeepSeek-VL~\cite{100} in a supervised manner, on the datasets mentioned above. The performance of these vision–language models was very low on this task, and the gap compared to other state-of-the-art methods was substantial. For this reason, we do not include their results in the main experiments. However, to demonstrate the generalizability of our proposed approach, we implemented our adaptive learning method on these models. With this integration, we were able to achieve significantly better performance. These results are presented in Section \ref{sec:generalization}. For the evaluation of risk analysis, we use the AUROC metric; for adaptive learning, we report the F1 score both per class and as a weighted average, along with AUC.

\subsection{Evaluation Results of Risk Analysis} \label{sec:RiskAnalysis}

\begin{figure*}
    \centering
    \resizebox{\textwidth}{!}{%
        \subfloat[BRACS (Original)]{\includegraphics[width=0.45\columnwidth]{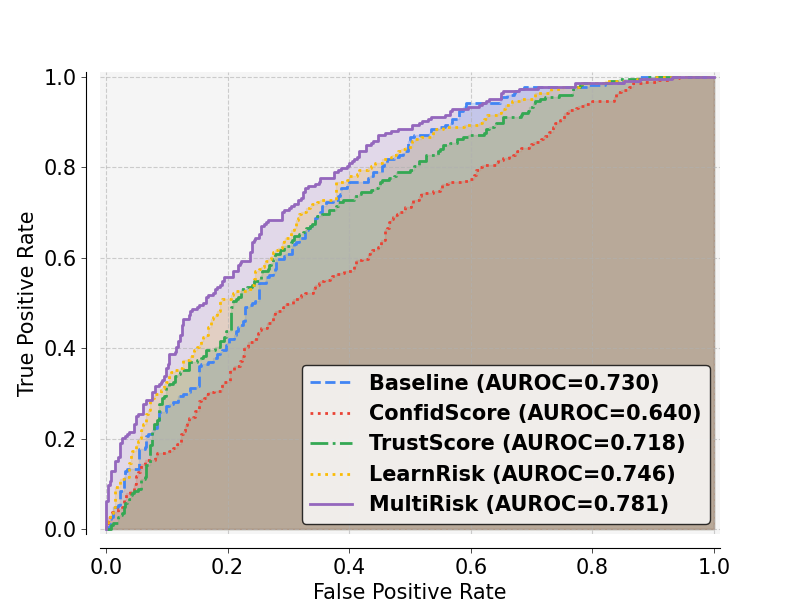}}
        \hspace{0.5cm}
        \subfloat[BRACS (512 × 512)]{\includegraphics[width=0.45\columnwidth]{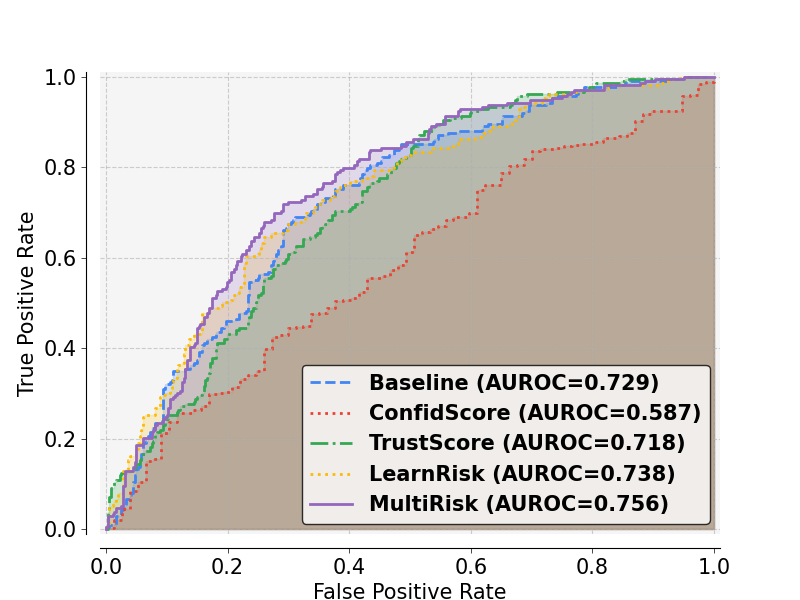}}
        \subfloat[BRACS-BACH]{\includegraphics[width=0.45\columnwidth]{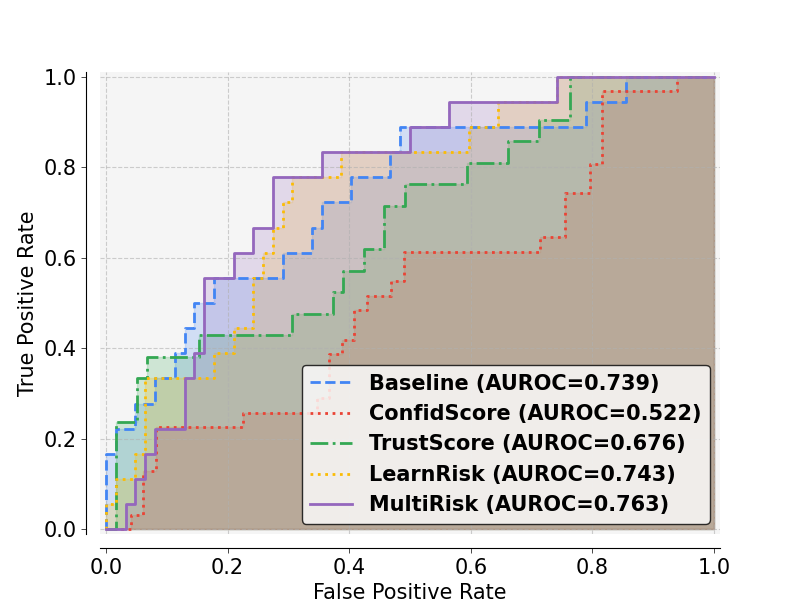}}
    }
    \caption{Performance of Risk Analysis on (a) Original BRACS, (b) BRACS(512×512), and (c) BRACS-BACH datasets.}
    \label{fig:roccomp}
\end{figure*}

Figure~\ref{fig:roccomp}  presents the detailed comparative results for AUROC analysis of MultRisk with SOTA methods On Original BRACS dataset, BRACS(512 x 512) variant, and the BRACS-BACH domain transfer dataset. It can be observed that on the original BRACS dataset in figure~\ref{fig:roccomp} (a), the AUROC of the Baseline stands at 73\% and the LearnRisk with 74.6\%, while ConfidNet, and TrustScore fails to surpass the Baseline's performance. Remarkably, MultiRisk demonstrates considerable improvement with an AUROC of 78.1\%, marking a noteworthy enhancement of 5.1\% compared to the Baseline. The results on the BRACS(512 x 512) dataset in figure~\ref{fig:roccomp} (b) are similar. The Baseline records an AUROC of 72.9\% and LearnRisk with 73.8\%, with ConfidNet, and TrustScore falling short of outperforming the Baseline in the multiclass setting. Once again, MultiRisk showcases its effectiveness by achieving an AUROC of 75.6\%, with an improvement of 2.7\% over the Baseline. Furthermore, for BRACS-BACH domain transfer four-class dataset, the baseline classifier was trained on BRACS samples but risk model was trained on BACH samples. The detailed evaluation in Figure~\ref{fig:roccomp} (c), It can be observed that MultiRisk achieved an AUROC of 76.3\%, a 2.4\% improvement over the Baseline's 73.9\%. While LearnRisk reached 74.3\%, both ConfidNet and TrustScore failed to outperform the Baseline. These results highlight MultiRisk's effectiveness in enhancing risk assessment models in the scenario of domain transfer. 

These results underscore the efficacy of the MultiRisk approach in enhancing risk analysis performance.

\subsection{Evaluation Results of Risk-based Adaptive Learning} \label{sec:AdaptiveLearning}

We present detailed comparative results in Table~\ref{tab:three_datasets}. On both the original BRACS and BRACS (512×512) datasets, results are reported for backbone methods, state-of-the-art (SOTA) breast cancer subtype and histopathological image prediction methods, as well as uncertainty estimation-based Learning techniques. For the domain transfer task, three SOTA domain adaptation methods are included. Across all three datasets and four categories of SOTA models, MultiRisk consistently outperforms the alternatives. By enhancing subtype prediction accuracy through misprediction risk analysis applied on baselines such as DenseNet121 and EfficientNetB4, MultiRisk demonstrates robustness and adaptability across datasets.

As shown in Table~\ref{tab:three_datasets}, MultiRisk achieves an F1-score of 61.15\% and an AUC of 91.38\% on the original BRACS dataset. Similarly, on the BRACS (512×512) dataset, it attains an F1-score of 65.98\% and an AUC of 92.07\%. While some methods report comparable results, MultiRisk consistently secures the highest metrics, highlighting its robustness and efficacy across different image resolutions. Furthermore, for the domain transfer task (BRACS-BACH), MultiRisk outperforms alternative approaches such as TSA, SCDA, and DANN in terms of both F1-score and AUC, underscoring its effectiveness in domain-transfer scenarios.

\begin{table*}[!ht]
\centering
\small 
\caption{Performance comparison (F1-score \% \& AUC \%) on BRACS (original), BRACS(512x512), and BRACS→BACH (domain shift). Best results are presented in bold. (mean and std across 5 runs)}
\label{tab:three_datasets}
\begin{tabular}{l|cc|cc|cc}
\hline
\multirow{2}{*}{Method} & \multicolumn{2}{c|}{\textbf{BRACS}} & \multicolumn{2}{c|}{\textbf{BRACS(512x512))}} & \multicolumn{2}{c}{\textbf{BRACS$\rightarrow$BACH}} \\
 & F1 & AUC & F1 & AUC & F1 & AUC \\
\hline
DenseNet-121      & 60.65$\pm$4.2 & 90.84$\pm$3.1 & 63.41$\pm$1.5 & 90.73$\pm$2.7 & 73.77$\pm$1.0    & 88.58$\pm$3.3 \\
ResNet-50         & 55.88$\pm$5.1 & 87.30$\pm$4.4 & 62.88$\pm$1.4 & 89.64$\pm$3.5 & 67.33$\pm$1.9    & 87.65$\pm$4.1 \\
EfficientNet-B4  & 59.39$\pm$4.7 & 88.29$\pm$3.7 & 56.74$\pm$1.9    & 89.17$\pm$4.5    &  78.04$\pm$1.0    & 92.35$\pm$2.8 \\
DenseNet-201     &   57.16$\pm$4.3   & 87.92$\pm$3.9   & 62.97$\pm$3.3 & 89.98$\pm$2.8 & 71.64$\pm$2.9    & 87.61$\pm$4.3\\
CCT              & 53.97$\pm$5.6 & 87.19$\pm$4.0 & 60.27$\pm$1.7 & 89.13$\pm$3.1 & 70.21$\pm$1.5    & 89.38$\pm$3.6 \\
TransPath        & 55.10$\pm$5.3 & 86.10$\pm$4.7 & 57.10$\pm$2.1 & 87.23$\pm$3.6 & 68.93$\pm$4.2    & 87.96$\pm$9.3 \\
CTransPath       & 56.75$\pm$5.0 & 87.14$\pm$3.9 & 59.70$\pm$1.1 & 88.70$\pm$3.5 & 70.98$\pm$3.9    & 89.04$\pm$6.7 \\

TransMIL       & 56.89$\pm$5.1 & 87.09$\pm$7.2 & 61.02$\pm$4.3 & 89.70$\pm$5.4 & 71.67$\pm$7.2    & 90.01$\pm$3.2 \\
CLAM       & 59.15$\pm$3.5 & 89.03$\pm$6.1 & 62.04$\pm$5.2 & 89.56$\pm$2.3 & 74.71$\pm$3.4    & 91.54$\pm$3.4 \\
SCDA             & --    & --    & --    & --    & 78.69$\pm$1.0   & 89.74$\pm$3.5 \\
TSA              & --    & --    & --    & --    & 72.01$\pm$1.3    & 86.76$\pm$4.4 \\
DANN             & --    & --    & --    & --    &  73.32$\pm$4.3    & 85.88$\pm$2.5\\
Deep Ensemble         & 58.36$\pm$4.8 & 89.63$\pm$3.5 & 62.70$\pm$1.9 & 90.74$\pm$3.3 & 70.84$\pm$1.5    & 89.54$\pm$3.4 \\
MCDropout         & 56.13$\pm$4.3 & 87.93$\pm$1.4 & 63.05$\pm$3.4 & 89.96$\pm$3.8 & 69.14$\pm$6.1    & 89.15$\pm$6.2 \\

LearnRisk        & 60.05$\pm$4.5 & 89.30$\pm$3.3 & 64.23$\pm$0.9 & 90.10$\pm$2.9 & 77.98$\pm$1.0    & 92.93$\pm$2.6 \\
MultiRisk (Ours) & \textbf{61.15$\pm$0.4} & \textbf{91.38$\pm$3.0} & \textbf{65.98$\pm$1.36} & \textbf{92.07$\pm$2.7} & \textbf{80.53$\pm$0.9}    & \textbf{93.23$\pm$2.5} \\
\hline
\end{tabular}
\end{table*}

Furthermore, BRACS is considered the benchmark data for breast cancer subtype prediction for which classwise F1 scores with weighted F1 are presented in state-of-the-art breast cancer subtype prediction methods like HACT-net, and ScoreNet. We have also presented the comparative evaluation results of classwise F1-score and weighted F1-score in Table~\ref{tab:f1scoreseven} for the backbone methods, state-of-the-art (SOTA) breast cancer subtype and histopathological image prediction methods, as well as uncertainty estimation-based learning techniques. It can be observed that MultiRisk achieves an impressive weighted F1 score of 65.98, surpassing the previous leading transformer-based approach, ScoreNet. Although ScoreNet excels in the flat epithelial atypia class, classical models such as Densenet-121, Densenet-201, and Resnet-50 outperform it in Atypical Ductal Hyperplasia, Usual Ductal Hyperplasia, and the histpathological image prediction method TransMIL outperforms it in Ductal Carcinoma in situ. Notably, MultiRisk improves the F1 score for 6 out of 7 classes considering the baseline in the subtype setting, demonstrating superior performance in the normal, pathological benign, and Invasive Carcinoma classes compared to the baseline. Specifically, it shows a 6.2\% improvement in normal ductal hyperplasia, and 4.41\% in flat epithelial atypia. For cancer classes, MultiRisk achieves notable enhancements of 1.6\% for ductal carcinoma in situ and 6.73\% for invasive carcinoma.

Despite not surpassing SOTA models in the Atypical Ductal Hyperplasia subtype, this can be attributed to the overall under-performance of non-baseline models in that category. HactNet is widely regarded as the most reliable approach for the BRACS dataset, validated against assessments by real-world pathologists. Notably, MultiRisk outperforms HactNet in nearly every class, achieving an overall improvement of 4.48\% in the weighted F1 score.
\begin{table*}[!htbp]
  \centering
  \caption{Mean and Standard Deviation evaluation for each class F1 and weighted F1 score using state-of-the-art methods for Benchmark BRACS Dataset(mean and std across 5 runs)}
  \label{tab:f1scoreseven}
  \small
  \setlength{\tabcolsep}{1pt}
  \begin{tabular}{l*{8}{c}}
    \hline
    \textbf{Method} & \textbf{0\_N} & \textbf{1\_PB} & \textbf{2\_UDH} & \textbf{3\_FEA} & \textbf{4\_ADH} & \textbf{5\_DCIS} & \textbf{6\_IC}  & \textbf{Weighted F1} \\ \hline
    Densenet-121 & 77.01$\pm$4.5 & 52.9$\pm$2.6 & 42.2$\pm$1.9 & 73.5$\pm$3.5 & \textbf{50.9$\pm$2.5} & 62.06$\pm$3.1 & 84.5$\pm$3.5 & 63.41$\pm$1.5 \\
    Densenet-201 & 77.83$\pm$3.9 & 52.3$\pm$2.3 & \textbf{57.6$\pm$2.7} & 69.2$\pm$3.3 & 44.7$\pm$2.0 & 58.46$\pm$4.2 & 80.7$\pm$3.6 & 62.97$\pm$3.1 \\
    Resnet-50 & 76.5$\pm$3.7 & 46.3$\pm$2.5 & 52.1$\pm$2.6 & 74.6$\pm$3.5 & 36.5$\pm$1.6 & 66.6$\pm$3.2 & 85.8$\pm$3.7 & 62.88$\pm$1.4 \\
    EfficientNet-B4 & 62.5$\pm$3.4 & 50.3$\pm$2.4 & 51.0$\pm$2.7 & 58.2$\pm$3.3 & 37.5$\pm$1.8 & 67.7$\pm$3.0 & 70.0$\pm$2.9 & 56.74$\pm$1.9\\

    CCT & 73.0$\pm$4.1 & 40.6$\pm$2.2 & 50.0$\pm$2.4 & 75.2$\pm$3.5 & 37.2$\pm$1.7 & 58.7$\pm$2.8 & 85.5$\pm$3.6 & 60.27$\pm$1.7 \\

    TransPath & 58.8$\pm$2.7 & 43.4$\pm$2.3 & 35.2$\pm$4.3 & 67.2$\pm$6.1 & 38.6$\pm$1.8 & 61.7$\pm$1.4 & 85.4$\pm$1.6 & 57.1$\pm$2.1 \\
    CTransPath & 60.0$\pm$1.4 & 47.1$\pm$2.7 & 37.9$\pm$2.3 & 72.9$\pm$3.1 & 43.3$\pm$3.8 & 65.9$\pm$1.4 & 91.0$\pm$2.6 & 59.7$\pm$1.1 \\

    TransMIL & 62.8$\pm$1.5 & 48.3$\pm$2.6 & 38.4$\pm$2.4 & 76.4$\pm$3.3 & 44.9$\pm$3.7 & \textbf{68.34$\pm$1.5} & 88.0$\pm$2.5 & 61.02±4.3  \\

     CLAM & 62.3$\pm$1.5 & 50.9$\pm$2.4 & 44.9$\pm$2.0 & 74.9$\pm$3.1 & 49.0$\pm$3.3 & 66.3$\pm$1.8 & 87.9$\pm$2.5& 62.04$\pm$5.3  \\
    HACT-Net & 61.6$\pm$2.1 & 47.5$\pm$2.9 & 43.6$\pm$1.9 & 74.2$\pm$1.4 & 40.4$\pm$2.5 & 66.4$\pm$2.6 & 88.4$\pm$0.2 & 61.5$\pm$0.9 \\
    ScoreNet & 64.3$\pm$1.5 & 54.0$\pm$2.2 & 45.3$\pm$3.4 & \textbf{78.1$\pm$2.8} & 46.7$\pm$1.0 & 62.9$\pm$2.0 & 91.0$\pm$1.4 & 64.4$\pm$0.9 \\
    
    Deep Ensemble & 75.7$\pm$3.3 & 47.7$\pm$3.0 & 49.1$\pm$1.8 & 77.8$\pm$2.8 & 39.1$\pm$1.8 & 60.86$\pm$2.7 & 87.34$\pm$3.1 & 62.70$\pm$1.9 \\
    MC Dropout(R50) & 77.3$\pm$3.4 & 54.5$\pm$2.6 & 56.9$\pm$7.1 & 70.0$\pm$3.9 & 50.8$\pm$2.8 & 58.1$\pm$6.3 & 73.7$\pm$5.3 & 63.05$\pm$3.6\\
    LearnRisk & 77.71$\pm$4.4 & 53.98$\pm$2.6 & 46.97$\pm$2.1 & 76.92$\pm$3.1 & 44.11$\pm$2.6 & 61.45$\pm$1.8 & 88.46$\pm$1.3 & 64.23$\pm$0.9 \\
    MultiRisk(Ours) & \textbf{78.30$\pm$4.3} & \textbf{54.71$\pm$2.6} & 48.40$\pm$2.3 & 77.91$\pm$3.9 & 47.62$\pm$2.1 & 63.66$\pm$2.1 & \textbf{91.23$\pm$2.7} & \textbf{65.98$\pm$1.0} \\
    \hline
  \end{tabular}
\end{table*}

\begin{figure*}
	\centering
    \resizebox{\textwidth}{!}{%
	\subfloat[BRACS(Original)]{\includegraphics[width=0.45\columnwidth]{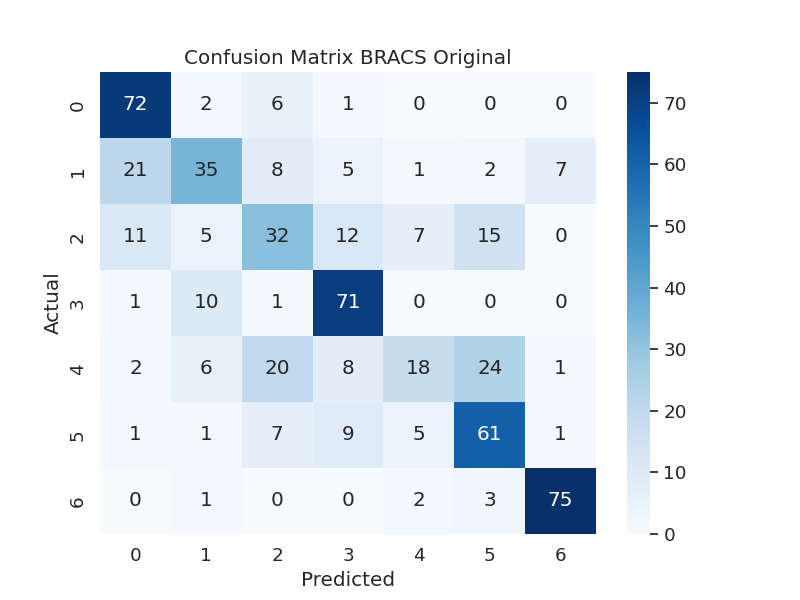}}\hspace{0.5pt}
    \subfloat[BRACS(512 x 512)]{\includegraphics[width=0.45\columnwidth]{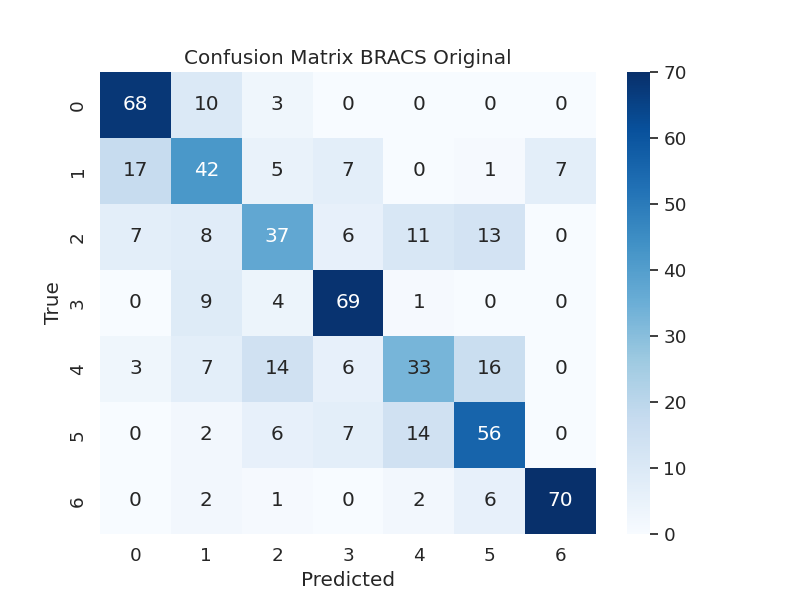}}\hspace{0.5pt}
	\subfloat[BRACS-BACH]{\includegraphics[width=0.45\columnwidth]{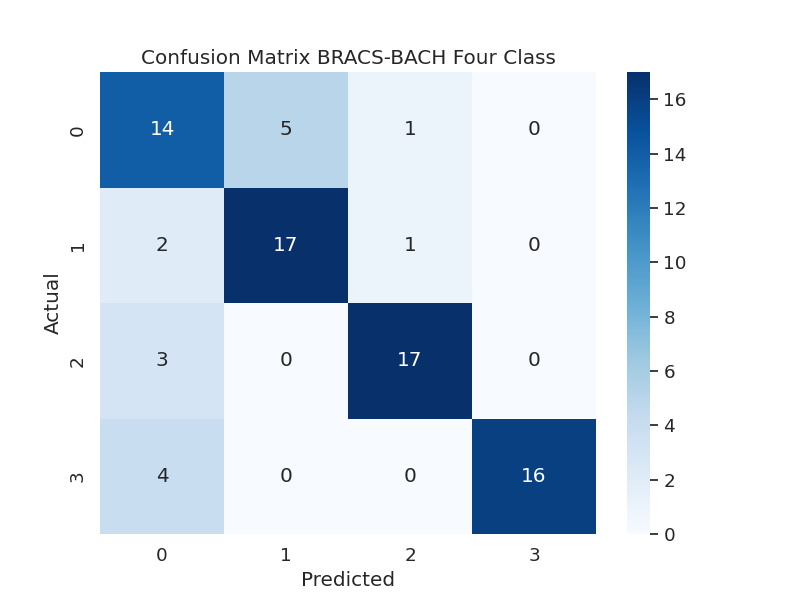}}\hspace{0.5pt}
    }
    \caption{Confusion Matrix Analysis of MultiRisk for all three datasets}
    \label{fig:confMatrix}
\end{figure*}

We have also visualized the comparative results by confusion matrix as shown in Figure \ref{fig:confMatrix}.
In Figure \ref{fig:confMatrix} (a), it can be observed that MultiRisk correctly predicts 364 of 570 test images on the original BRACS dataset, achieving the highest accuracy in "Invasive Carcinoma" (75/81), while mispredictions mainly occurred between "Pathological Benign," "Usual Ductal Hyperplasia," and "Atypical Ductal Hyperplasia." In Figure \ref{fig:confMatrix} (b), for the BRACS(512 x 512) dataset, MultiRisk accurately predicts 375 of 512 images, with similar errors in distinguishing between "Pathological Benign," "Usual Ductal Hyperplasia," and "Atypical Ductal Hyperplasia." Finally, for the domain transfer analysis, the confusion matrix in Figure \ref{fig:confMatrix} (c) shows that, using 80 test images from BRACS-BACH, MultiRisk accurately classifies 64 images, with mispredictions mainly between "Normal" and "Benign" classes, highlighting both strengths and areas for improvement in class differentiation and domain transfer.

\begin{figure}
  \centering
  \includegraphics[width=\columnwidth]{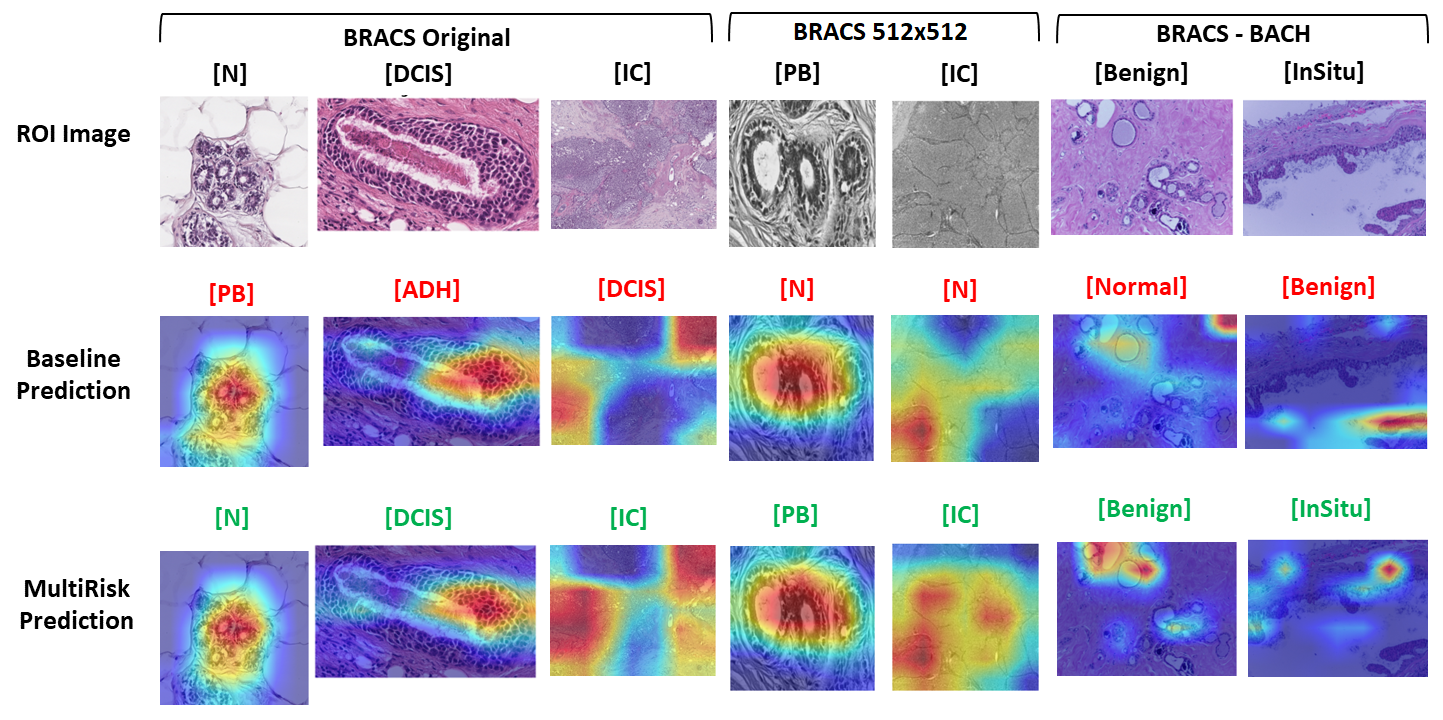}
  \caption{Qualitative Comparison between Baseline and MultiRisk, class labels and predictions, where red denotes misprediction and green denotes correctly predicted}
  \label{fig:qualitative}
\end{figure}

Finally, to analyze the qualitative performance of MultiRisk-based adaptive training, we showcase a qualitative analysis of ROI images across all three datasets using GradCam~\cite{86}, which generates heatmaps of class activations, in Figure~\ref{fig:qualitative}. The results indicate that MultiRisk not only corrects mispredictions for all classes but also identifies additional tumor regions. Notable corrections include transitions from ADH to DCIS, DCIS to IC in the original BRACS dataset, N to IC in BRACS(512x512), and Benign to InSitu during the domain transfer task.

\subsection{Sensitivity Analysis} \label{sec:Sensitivity}

  In the sensitivity analysis, we have evaluated the sensitivity of risk model w.r.t validation samples by conducting two experiments, the first experiment using a different number of validation samples, and the second one using the same number but randomly selected validation samples. Note that the training and test samples remain the same in the sensitivity study. Additionally, we have also evaluated the efficacy of MultiRisk by leveraging different baseline prediction models.


\begin{figure*}
    \centering
    \resizebox{\textwidth}{!}{%
    \subfloat[BRACS(Original)]{\includegraphics[width=0.45\textwidth]{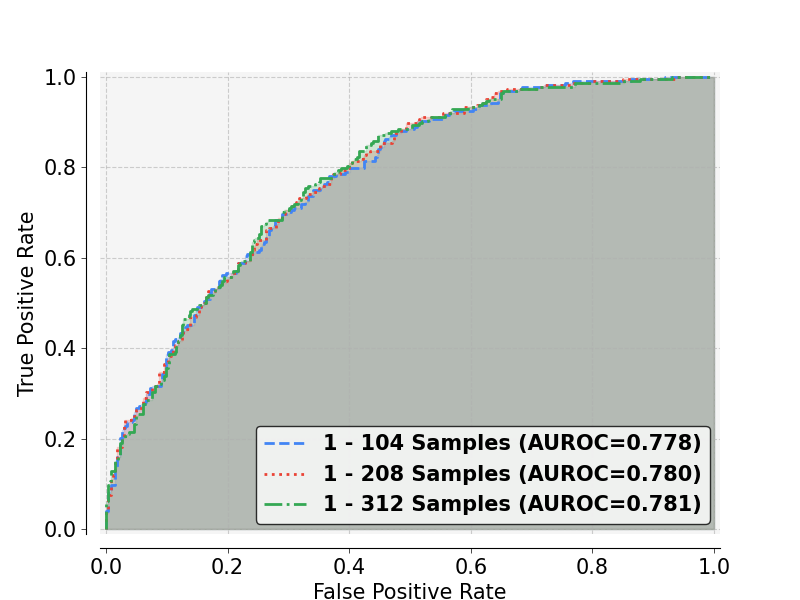}}\hspace{0.5pt}
    \subfloat[BRACS(512 x 512)]{\includegraphics[width=0.45\textwidth]{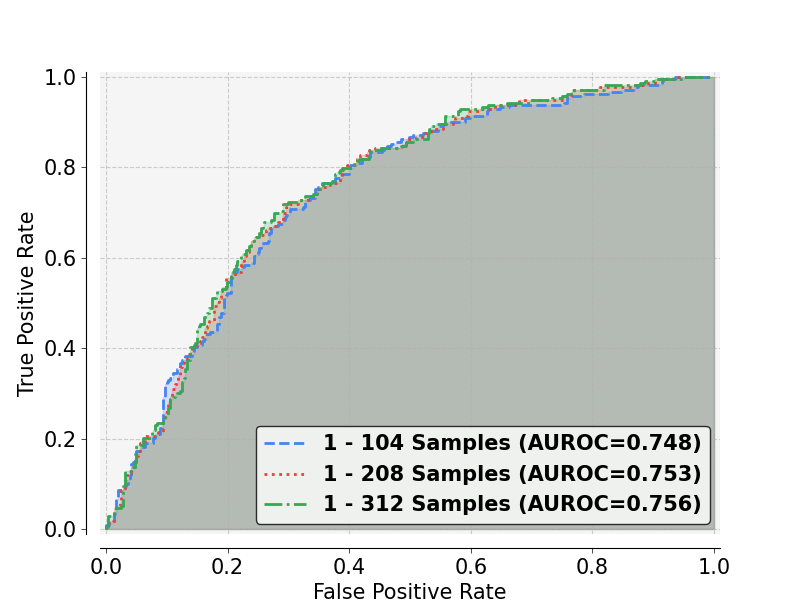}}\hspace{0.5pt}
    \subfloat[BRACS-BACH]{\includegraphics[width=0.45\textwidth]{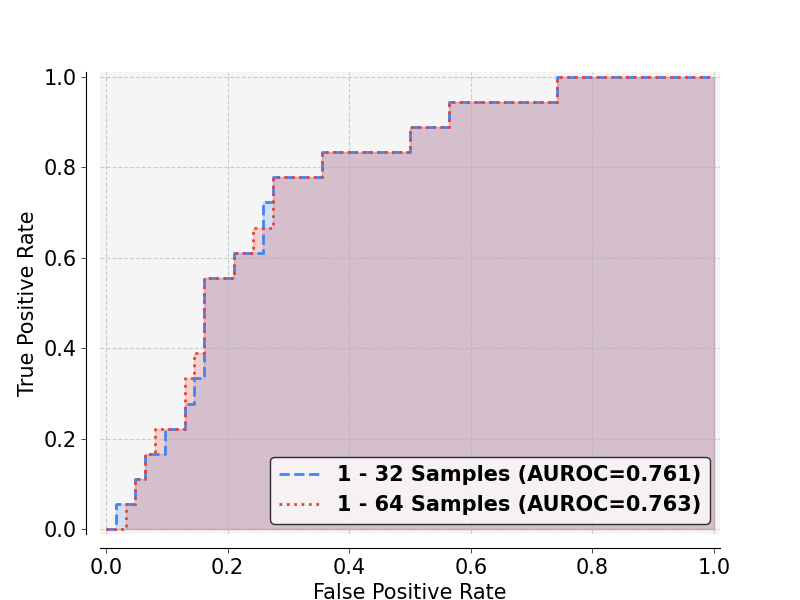}}
    }
    
    \vspace{-10pt}
    
    \resizebox{\textwidth}{!}{%
    \subfloat[BRACS(Original)]{\includegraphics[width=0.45\textwidth]{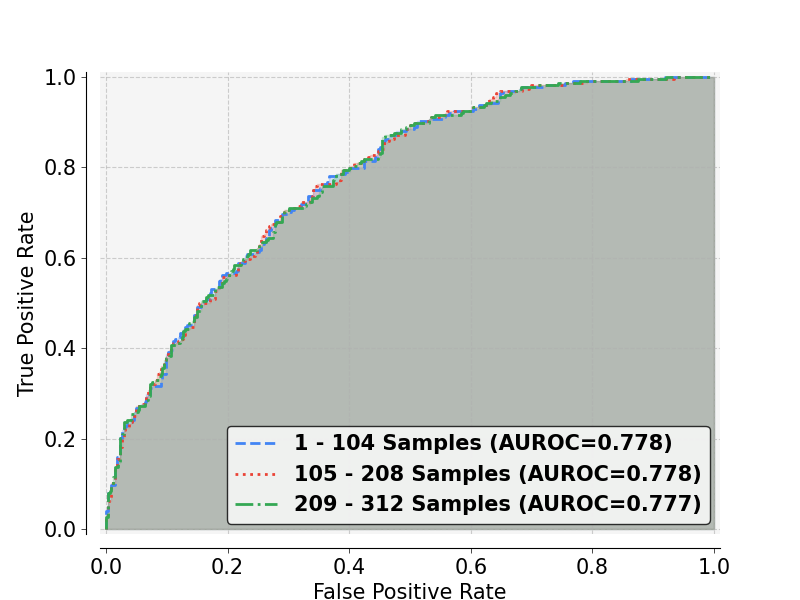}}\hspace{0.5pt}
    \subfloat[BRACS(512 x 512)]{\includegraphics[width=0.45\textwidth]{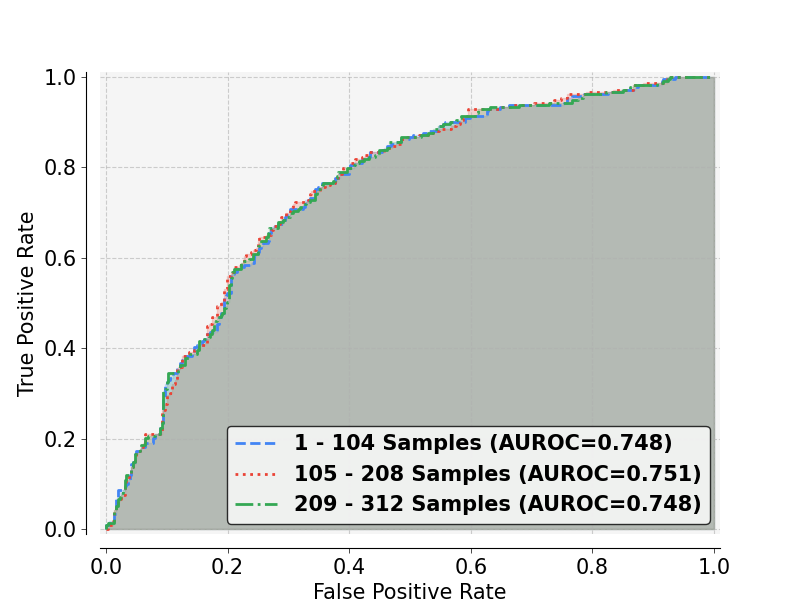}}\hspace{0.5pt}
    \subfloat[BRACS-BACH]{\includegraphics[width=0.45\textwidth]{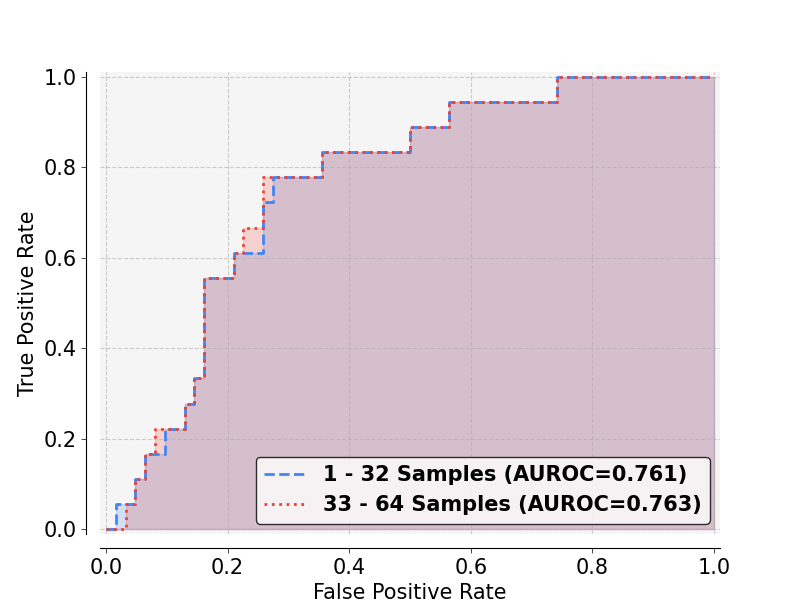}}
    }
    
    \caption{Sensitivity Analysis of MultiRisk for all three datasets, Validation Samples with Different number and type of samples varying from 104 to 312, and 0, 312}
    \label{fig:samples_number}
    
    \vspace{-5pt}
    
\end{figure*}

The evaluation results w.r.t the number of validation samples have been presented in Figure~\ref{fig:samples_number} (a), (b) and (c). The evaluation results w.r.t random validation samples have been presented in Figure~\ref{fig:samples_number} (d), (e) and (f). It can be observed that in both cases, the performance of MultiRisk is robust across all three datasets. A nominal change under 1\% is recorded while performing these experiment across all three datasets.

Furthermore, the evaluation results on the performance sensitivity of MultiRisk-based adaptive learning with different Baseline models have been presented in Table~\ref{tab:model_comparison}. It can be observed that on each baseline model, MultiRisk outperforms both the baseline and LearnRisk in terms of f1 and AUC. These results clearly demonstrate the performance robustness of the MultiRisk-based approach. 

\begin{table}[htbp]
\centering
\caption{Performance Comparison of MultiRisk Based on Different Baselines}
\label{tab:model_comparison}
\footnotesize 
\setlength{\tabcolsep}{2.5pt} 
\begin{tabular}{lcccccc}
\toprule
Model & \multicolumn{2}{c}{Baseline} & \multicolumn{2}{c}{LearnRisk} & \multicolumn{2}{c}{MultiRisk} \\
 & F1 & AUC & F1 & AUC & F1 & AUC \\
\midrule
Resnet50 & 62.88 & 89.64 & 63.40 & 88.93 & \textbf{64.08} & \textbf{91.04} \\
EfficientNetB4 & 56.74 & 89.17 & 56.36 & 89.95 & \textbf{57.03} & \textbf{91.49} \\
CCT & 60.27 & 89.13 & 60.27 & 90.61 & \textbf{61.97} & \textbf{92.03} \\
Densenet201 & 62.97 & 89.98 & 63.03 & 90.31 & \textbf{63.10} & \textbf{91.48} \\
Densenet121 & 63.44 & 90.73 & 64.23 & 90.10 & \textbf{65.98} & \textbf{92.07} \\
\bottomrule
\end{tabular}
\end{table}

\subsection{Ablation Study}
\label{sec:ablation}

  To validate the design choices in Multiclass misprediction risk analysis and the MultiRisk-based adaptive learning framework, we conducted an ablation study comparing four key components. First, for risk feature generation, we evaluated the impact of using features from each model without feature selection and fusion. Second, to assess the Risk Model, we compared our attention-based weighting mechanism with traditional Gaussian weighting. Third, we evaluated our voting-based learning-to-rank strategy against conventional pairwise comparison-based ranking. Finally, to examine the handling of class imbalance, we compared our approach with established methods such as Focal Loss and Class-Balanced Loss. For feature extraction. attention based weighting, and learning-to-rank experiments, F1 scores were recorded for adaptive learning and Risk-AUROC for risk analysis. For class imbalance assessment, both F1 and AUC metrics were computed. Across all scenarios, MultiRisk consistently outperformed alternative configurations. The results are summarized in Table~\ref{tab:ablation_table}.

\begin{table}[htbp]
\scriptsize
\centering

\setlength{\tabcolsep}{4pt}
\caption{Ablation Study: Comparative Analysis of Our Design Choices and Innovations with Traditional Methods}
\label{tab:ablation_table}

\begin{tabular}{l
                cc
                cc
                cc}
\toprule
\multicolumn{7}{c}{\textbf{Feature Extraction}} \\
\midrule
Method & \multicolumn{2}{c}{BRACS} & \multicolumn{2}{c}{BRACS-512} & \multicolumn{2}{c}{BRACS-BACH} \\
\cmidrule(lr){2-3} \cmidrule(lr){4-5} \cmidrule(lr){6-7}
       & F1 (\%) & Risk-AUROC (\%) & F1 (\%) & Risk-AUROC (\%) & F1 (\%) & Risk-AUROC (\%) \\
\midrule
Resnet50 & 59.03 & 70.1 & 62.35 & 69.2 & 76.12 & 65.8 \\
Densenet121  & 60.86 & 72.2 & 64.81 & 73.9 & 76.56 & 67.4 \\
EfficientnetB4  & 59.87 & 65.7 & 62.29 & 70.2 & 77.86 & 73.3 \\
CCT  & 53.93 & 70.4 & 62.18 & 74.7 & 72.81 & 72.3 \\
\textbf{MultiRisk(Ours)} & \textbf{61.15} & \textbf{78.1} &\textbf{65.98} &\textbf{75.6} & \textbf{80.53} & \textbf{75.3} \\
\midrule
\multicolumn{7}{c}{\textbf{Risk Model Weights}} \\
\midrule
Method & \multicolumn{2}{c}{BRACS} & \multicolumn{2}{c}{BRACS-512} & \multicolumn{2}{c}{BRACS-BACH} \\
\cmidrule(lr){2-3} \cmidrule(lr){4-5} \cmidrule(lr){6-7}
       & F1 (\%) & Risk-AUROC (\%) & F1 (\%) & Risk-AUROC (\%) & F1 (\%) & Risk-AUROC (\%) \\
\midrule
Gaussian Weights & 60.20  & 78.1 & 64.23 & 75.63& 76.12 & 72.0 \\

\textbf{MultiRisk(Ours)} & \textbf{61.15} & \textbf{78.1} &\textbf{65.98} &\textbf{75.6} & \textbf{80.53} & \textbf{75.3} \\

\midrule
\multicolumn{7}{c}{\textbf{Learning to Rank}} \\
\midrule
Method & \multicolumn{2}{c}{BRACS} & \multicolumn{2}{c}{BRACS-512} & \multicolumn{2}{c}{BRACS-BACH} \\
\cmidrule(lr){2-3} \cmidrule(lr){4-5} \cmidrule(lr){6-7}
       & F1 (\%) & Risk-AUROC (\%) & F1 (\%) & Risk-AUROC (\%) & F1 (\%) & Risk-AUROC (\%) \\
\midrule
Pairwise Comparison & 60.05 & 76.4 & 64.23 & 73.8 & 77.98    & 74.3 \\ 
\textbf{Voting Based(Ours)} & \textbf{61.15} & \textbf{78.1} &\textbf{65.98} &\textbf{75.6} & \textbf{80.53} & \textbf{75.3} \\

\midrule
\multicolumn{7}{c}{\textbf{Class Imbalance}} \\
\midrule
Method & \multicolumn{2}{c}{BRACS} & \multicolumn{2}{c}{BRACS-512} & \multicolumn{2}{c}{BRACS-BACH} \\
\cmidrule(lr){2-3} \cmidrule(lr){4-5} \cmidrule(lr){6-7}
       & F1 (\%) & AUC (\%) & F1 (\%) & AUC (\%) & F1 (\%) & AUC (\%) \\
\midrule
Focal Loss & 60.92 & 91.82 & 64.37 & 91.37 & 76.75 & 88.03 \\
Class-Balanced Loss & 58.92 & 89.62 & 62.35 & 90.12 & 78.69 & 90.85 \\
\textbf{MultiRisk(Ours)} & \textbf{61.15} & \textbf{91.38} &\textbf{65.98} &\textbf{92.07} & \textbf{80.53} & \textbf{93.23} \\
\bottomrule
\end{tabular}
\end{table}

\subsection{Generalization Analysis}
\label{sec:generalization}

We evaluated MultiRisk on two diverse lung-colon cancer LC25000~\cite{98}, and lung cancer LungHist700~\cite{98} histopathology datasets to test its generalizability beyond breast cancer. These results are presented in table \ref{tab:lung_generalization_compact}, compared to baseline model(Densenet121), MultiRisk consistently achieves higher F1 and AUC scores across almost all the data scenarios.

Notably, in domain transfer scenarios like training on LC25000 and testing on LungHist700 and vice versa, MultiRisk maintains superior performance, demonstrating strong adaptability. These results highlight MultiRisk’s effectiveness in improving prediction accuracy and robustness across different datasets and cancer types, supporting its broader clinical applicability.

    Furthermore, our misprediction risk analysis leverages features extracted from deep learning models, making MultiRisk readily adaptable to various neural network architectures, including complex multi-modal Vision-Language Models (VLMs). To demonstrate our method's generalization strength, we evaluated it on three supervised architectures: CLIP-style ResNet-50 with GPT-2, ResNet-50 with LLaVA-Med, and ResNet-50 with DeepSeek. As shown in Table \ref{tab:vlm_generalization}, MultiRisk delivered substantial performance gains over the baseline VLMs in all scenarios.

\begin{table}[htbp]
\centering
\footnotesize
\setlength{\tabcolsep}{8pt}
\caption{Generalization Results: Baseline vs. MultiRisk on LC25000 and LungHist700 Datasets}
\label{tab:lung_generalization_compact}

\begin{tabular}{lcc|cc}
\toprule
\multirow{2}{*}{\textbf{Dataset}} 
& \multicolumn{2}{c|}{\textbf{Baseline}} 
& \multicolumn{2}{c}{\textbf{MultiRisk}} \\
\cmidrule(lr){2-3} \cmidrule(lr){4-5}
& F1 (\%) & AUC (\%) & F1 (\%) & AUC (\%) \\
\midrule
LC25000               & 96.34 & \textbf{99.53} & \textbf{98.19} & 99.24 \\
LungHist700            & 80.67 & 92.32 & \textbf{84.88} & \textbf{93.05} \\
LC25000 $\rightarrow$ LungHist700 & 68.43 & 90.71 & \textbf{72.64} & \textbf{92.62} \\
LungHist700 $\rightarrow$ LC25000 & 65.74 & 90.71 & \textbf{68.73} & \textbf{92.19} \\
\bottomrule
\end{tabular}

\end{table}

\begin{table}[htbp]
\centering
\footnotesize
\setlength{\tabcolsep}{8pt}
\caption{Generalization Results: Baseline vs. MultiRisk on Vision--Language Models (BRACS Dataset)}
\label{tab:vlm_generalization}

\begin{tabular}{lcc|cc}
\toprule
\multirow{2}{*}{\textbf{Vision--Language Model}} 
& \multicolumn{2}{c|}{\textbf{Baseline}} 
& \multicolumn{2}{c}{\textbf{MultiRisk}} \\
\cmidrule(lr){2-3} \cmidrule(lr){4-5}
& F1 (\%) & AUC (\%) & F1 (\%) & AUC (\%) \\
\midrule
CLIP(ResNet-50 + GPT2)     & 51.43 & 85.21 & \textbf{54.32} & \textbf{86.76} \\
CLIP(ResnNet-50 + LLaVA-Med)           & 52.11 & 85.98 & \textbf{54.78} & \textbf{87.85} \\
DeepSeek-VL(ResNet-50)     & 54.65 & 87.71 & \textbf{57.41} & \textbf{89.09} \\

\bottomrule
\end{tabular}

\end{table}

\subsection{Computational Efficiency and Scalability Discussion}

\label{sec:scalability}

While MultiRisk achieves substantial improvements in prediction performance and robustness, we recognize the critical importance of computational efficiency and scalability in resource-constrained diagnostic settings. To address this, our adaptive learning process is deliberately constrained to a short training duration of only 10 epochs, significantly reducing the computational burden compared to standard prolonged training regimes. Furthermore, we validate MultiRisk on lower-resolution images (512×512), demonstrating competitive and better performance that aligns with the reduced memory and processing requirements inherent in such inputs, making the approach well-suited for deployment on hardware with limited resources.

Additionally, the misprediction risk analysis model is sample-efficient, requiring only a small subset of training samples to effectively calculate risk and guide adaptive learning. This targeted training avoids extensive overhead and contributes to the method’s scalability. Collectively, these design choices ensure that MultiRisk maintains a favorable balance between performance gains and computational practicality, positioning it as a viable solution for real-world clinical applications with constrained computational resources.
\FloatBarrier

\section{Conclusion and Future Work}

   In this paper, we proposed a novel approach to multiclass misprediction risk analysis, \textbf{MultiRisk}, for breast cancer subtype prediction. MultiRisk introduces new mechanisms for risk feature generation, risk model construction and training, and risk-based adaptive training, all specifically designed for multiclass problems to enable accurate misprediction quantification. Building on MultiRisk, we further presented an adaptive learning strategy capable of efficiently fine-tuning a baseline subtype prediction model to a target workload of breast cancer. Extensive experiments on real benchmark datasets demonstrate the effectiveness, robustness, and generalization ability of our approach across different types of cancer and diverse deep learning models.

    For future work, we note that the proposed \textbf{MultiRisk} framework is potentially applicable to other types of prediction tasks in healthcare; however, its technical details warrant further exploration. Additionally, the current solution derives risk metrics from mainstream deep neural network (DNN) models, which inherently have limited interpretability. To improve both the effectiveness and interpretability of risk analysis, future research could investigate generating risk features through explainable AI techniques.

\section*{Data Availability}

The datasets used in this study are publicly available and open source for research purpose. The BRACS: BReAst Carcinoma Sub-typing can be accessed at \url{https://www.bracs.icar.cnr.it/}. ICIAR 2018 Grand Challenge on Breast Cancer Histology Dataset i.e BACH at \url{https://iciar2018-challenge.grand-challenge.org/Dataset/}. Additionally, the dataset used for generalization experiments can be accessed at \url{https://zenodo.org/records/14998042} and \url{https://www.kaggle.com/datasets/abdullahhasansajjad/lunghist700}. The code for MultiRisk is available at \url{https://github.com/SheerazNWPU/MultiRISK}. Additional data or resources can be provided upon reasonable request.

\section*{Author Contributions}
Gul Sheeraz led the study, handled data curation, formal analysis, investigation, methodology development, project administration, and drafting. Qun Chen provided funding, resources, supervision, and manuscript review. Liu Feiyu contributed to analysis, investigation, methodology, software, validation, visualization, and manuscript review. Zhou Fengjin provided co-supervision related to medical information for this task.
\section*{Declaration of Generative AI and AI-assisted technologies in the writing process}
The authors used ChatGPT 4.0 to enhance writing flow in some sections. They reviewed and edited the content afterward, taking full responsibility for the final manuscript.
\section*{Acknowledgments}
This work is supported by the National Key Research and Development Program of China under the project number 2023YFB4503600, and National Natural Science Foundation of China under Grant No. 62172335.
\section*{Conflict of Interest}
The authors declare that they have no known competing financial interests or personal relationships that could have appeared to influence the work reported in this paper.


\bibliographystyle{cas-model2-names}

\bibliography{cas-refs}

\end{document}